\pdfoutput=1

\documentclass[11pt]{article}

\usepackage[final]{acl}

\usepackage{times}
\usepackage{latexsym}
\usepackage{longtable}
\usepackage[T1]{fontenc}
\usepackage{listings}

\usepackage[utf8]{inputenc}

\usepackage{microtype}

\usepackage{inconsolata}

\usepackage{booktabs}
\usepackage{xspace}
\usepackage{graphicx}
\usepackage{float}
\usepackage{url}
\usepackage{adjustbox}
\usepackage{xcolor}
\usepackage{algpseudocode}
\usepackage{longtable}
\usepackage[most]{tcolorbox}
\usepackage{mdframed}
\usepackage{lipsum}
\usepackage{multirow}
\usepackage{amsfonts}
\usepackage{rotating}
\usepackage{enumitem}
\usepackage[capitalize]{cleveref}
\usepackage{colortbl}
\usepackage{tabularray}
\usepackage{makecell}
\usepackage{caption}
\usepackage{svg}

\newcommand*{\fixedimg}[1]{%
    \raisebox{-.05cm}{%
        \includegraphics[
        height=0.3cm,
        width=0.3cm,
        keepaspectratio,
        ]{#1}%
    }%
}

\newcommand{\eat}[1]{}

\newif\ifcomments

\ifcomments
  \newcommand{\showhidecomments}{%
    \newcommand{\tushar}[1]{\textcolor{red}{$_{TK}${[##1]}}}%
    \newcommand{\nb}[1]{\textcolor{blue}{$_{NB}${[##1]}}}%
    \newcommand{\ashish}[1]{\textcolor{brown}{$_{AS}${[##1]}}}%
    \newcommand{\harsh}[1]{\textcolor{teal}{$_{HT}${[##1]}}}%
    \newcommand{\mareike}[1]{\textcolor{green}{$_{MR}${[##1]}}}%
    \newcommand{\shashank}[1]{\textcolor{orange}{$_{SG}${[##1]}}}%
  }
\else
  \newcommand{\showhidecomments}{%
    \newcommand{\tushar}[1]{}%
    \newcommand{\nb}[1]{}%
    \newcommand{\ashish}[1]{}%
    \newcommand{\harsh}[1]{}%
    \newcommand{\mareike}[1]{}%
    \newcommand{\shashank}[1]{}%
  }
\fi
\showhidecomments

\tcbset{
    colback=white,
    colframe=red!75!black,
    width=\textwidth,
    arc=0mm,
    boxrule=0.5mm,
}
 
\crefname{appsec}{Apdx.}{Apdx.}
\Crefname{appendix}{App.}{Apps.}
\crefformat{section}{\S#2#1#3}
\crefformat{subsection}{\S#2#1#3}
\crefformat{subsubsection}{\S#2#1#3}

\title{
VehicleWorld: A Highly Integrated Multi-Device Environment for Intelligent Vehicle Interaction
}

\author{
Jie Yang\thanks{Equal contribution.} \quad
Jiajun Chen$^*$ \quad
Zhangyue Yin$^*$ \quad
Shuo Chen \quad
Yuxin Wang \quad
\\
\textbf{
Yiran Guo \quad
Yuan Li \quad
Yining Zheng \textsuperscript{\dag} \quad
Xuanjing Huang \quad
Xipeng Qiu\textsuperscript{\dag}
}
\\
School of Computer Science and Artificial Intelligence, Fudan University \\
\texttt{\{yangj24,jiajunchen22,yinzy21,shuochen24\}@m.fudan.edu.cn} \\
\texttt{\{wangyuxin21,yrguo23,liyuan24\}@m.fudan.edu.cn}\\
\texttt{\{ynzheng19,xjhuang,xpqiu\}@fudan.edu.cn} \quad
}

\begin{document}
\maketitle

\begin{abstract}
Intelligent vehicle cockpits present unique challenges for API Agents, requiring coordination across tightly-coupled subsystems that exceed typical task environments' complexity. 
Traditional Function Calling (FC) approaches operate statelessly, requiring multiple exploratory calls to build environmental awareness before execution, leading to inefficiency and limited error recovery.
We introduce VehicleWorld, the first comprehensive environment for the automotive domain, featuring 30 modules, 250 APIs, and 680 properties with fully executable implementations that provide real-time state information during agent execution.
This environment enables precise evaluation of vehicle agent behaviors across diverse, challenging scenarios.
Through systematic analysis, we discovered that direct state prediction outperforms function calling for environmental control.
Building on this insight, we propose State-based Function Call (SFC), a novel approach that maintains explicit system state awareness and implements direct state transitions to achieve target conditions.
Experimental results demonstrate that SFC significantly outperforms traditional FC approaches, achieving superior execution accuracy and reduced latency.
We have made all implementation code publicly available on GitHub\footnote{\url{https://github.com/OpenMOSS/VehicleWorld}}.
\end{abstract}
\begin{figure}[t]
  \centering
  \includegraphics[width=\linewidth]{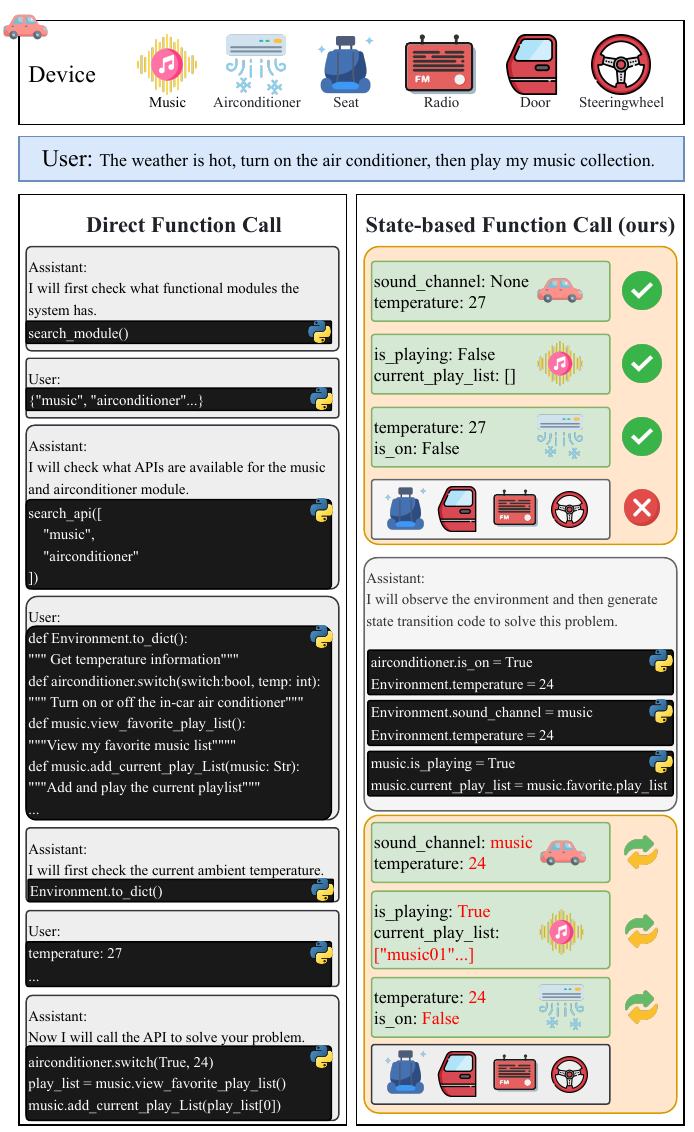}
  \caption{Direct Function Call (FC) versus State-based Function Call (SFC)}
  \vspace{-.5em}
  \label{fig:intro}
\end{figure}

\section{Introduction}
\label{sec:introduction}

API Agents represent a paradigm shift in intelligent interaction by combining large language models' cognitive capabilities with external tools' execution capabilities~\citep[][\emph{inter alia}]{durante2024agentaisurveyinghorizons,zhang2025igniting,qu2025tool}. These agents distinguish themselves through autonomous decision-making, sophisticated reasoning abilities, and seamless tool interaction, transcending traditional dialogue systems' limitations~\citep{xi2025rise,jin2024toolbridge,cao2024tool}.

Intelligent vehicle cockpits present a uniquely challenging domain for API Agents, integrating numerous tightly-coupled subsystems from entertainment and navigation to vehicle diagnostics and environmental controls. Within this environment, agents must orchestrate diverse systems while allowing drivers to maintain focus on the road~\citep{ma2024development}. Despite their importance to modern vehicle systems, the field lacks a comprehensive evaluation framework for these cockpit agents, preventing systematic assessment of their performance across implementations~\citep{khiabani2025optimizing}.

Figure~\ref{fig:intro} illustrates the challenges with a common request: ``The weather is hot, turn on the air conditioner, then play my music collection.'' This seemingly simple instruction requires coordination across multiple subsystems. Traditional Function Call (FC) approach operates statelessly, sequentially exploring available modules and APIs, necessitating multiple exploratory calls to build environmental awareness before execution. As noted by \citet{guo2024stabletoolbench}, this approach becomes problematic when API calls fail, as agents struggle to recover without a macroscopic understanding of the global state. Additionally, agents can only discern execution results through limited API return information, which may lead to incorrect conclusions about task success or failure.

To address these limitations, we developed VehicleWorld, a virtual intelligent cockpit environment supporting 30 modules, 250 APIs, and 680 properties. All APIs are executable code implementations, with each corresponding to a module instance method implemented through attribute state modifications. Based on our comprehensive environment construction, we discovered that state information significantly enhances agent call accuracy. We propose State-based Function Call (SFC), which explicitly constructs state transition processes by maintaining awareness of the system's current state and directly implementing necessary transitions to achieve target states. Our contributions include:
\begin{itemize}
\item The first comprehensive environment for the automotive domain that provides real-time state information during model execution.
\item A novel State-based Function Call (SFC) approach specifically engineered for our VehicleWorld environment.
\item Experiments demonstrate that SFC exhibits significant improvements in execution accuracy and latency reduction compared to FC.
\end{itemize}

\section{Related Work}
\label{sec:related_work}
\paragraph{Tool-Utility Agent}
Recent studies have demonstrated that integrating tool-using capabilities significantly enhances the adaptability and effectiveness of agents in complex environments~\citep{mialon2023augmented, schick2023toolformer}. Tools expand agents' operational range, enabling flexible interactions with dynamic environments~\citep{chen2024can, nakano2021webgpt, ma2024sciagent}, provide valuable feedback mechanisms for self-reflection~\citep{yu2024steptool, liu2024summary, wang2024llms}, and bridge knowledge gaps encountered by large language models~\citep{li2025search}. Vehicle scenarios present unique challenges at the intersection of real-time responsiveness, multimodal human-machine interaction, and strict safety requirements. Despite considerable advancements in tool-utility agents, research specifically tailored to intelligent vehicle cockpits remains limited. Modern vehicle cockpits have evolved into human-machine interactive systems~\citep{ma2024development}, underscoring the need for dedicated studies addressing their specialized design, user intentions, and evaluation methodologies.

\paragraph{Simulated World}
World models enable agents to build internal representations of environments, enhancing decision-making capabilities. These models can be categorized as model-based worlds~\citep{ha2018world} that represent environments through learned neural network parameters, or code-based worlds~\citep{tang2024worldcoder,trivedi2024appworld} that provide more interpretable representations by encoding environmental dynamics as executable code. While WorldCoder~\citep{tang2024worldcoder} constructs world models through Python programs derived from environmental interactions, and AppWorld~\citep{trivedi2024appworld} creates a simulated environment of applications with numerous APIs, these approaches have limitations for automotive cockpit environments: (1) Apps operate in isolation, lacking the systemic coupling necessary between modules; (2) AppWorld cannot explicitly represent real-time application states, limiting models' understanding of current system conditions. To address these limitations, VehicleWorld provides an executable intelligent cockpit environment with well-defined APIs and direct state observability, enabling agents to develop both procedural and declarative knowledge in automotive scenarios.

\paragraph{Function Call}
Function calling has emerged as a critical mechanism for transforming LLMs into effective tool-using agents. Recent advancements include specialized multi-task learning frameworks~\citep{abdelaziz2024granite}, data-centric approaches generating high-quality datasets~\citep{liu2024apigen, liu2024toolace}, parallel function execution paradigms~\citep{zhang2016mechanism}, and robust security frameworks~\citep{srinivasan2023nexusraven}. While these approaches have made significant progress, they predominantly focus on optimizing the function calling process itself, either by enhancing model instruction-following ability~\citep{chen2024agent} or simplifying complex function calling sequences~\citep{huang2024planning}. Our work takes a complementary direction by reconceptualizing tool use through a state-based framework, introducing a state-transparent environment where agents directly access and operate system states, enabling them to predict desired goals and generate efficient state transition code.

\section{VehicleWorld}
\label{sec:vehicle_world}

To simulate realistic intelligent cockpit environments, we introduce \textbf{VehicleWorld}, the first comprehensive environment for the automotive domain that provides real-time state information during agent execution and supports precise evaluation of vehicle agent behaviors.

\begin{figure*}[t]
	\raggedleft
	\centering
	\includegraphics[width=1.0\textwidth]{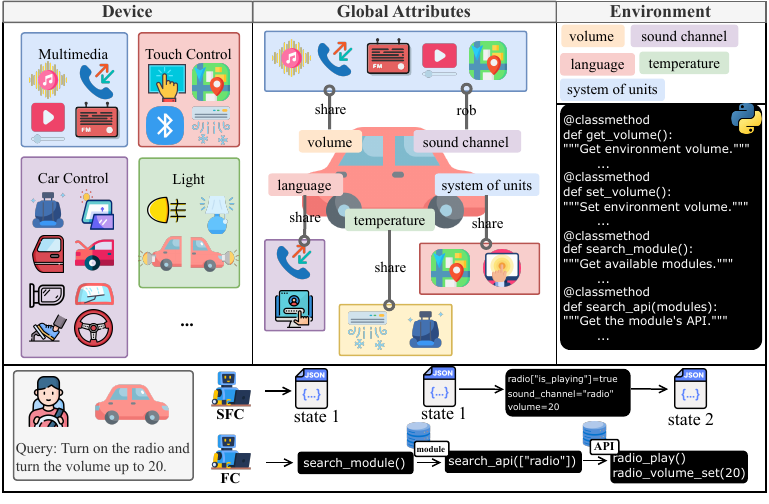}
	\caption{\textbf{Overview of VehicleWorld}. Above is the composition of VehicleWorld, which consists of 30 devices and 680 attributes. To maintain common attributes within the system, we have established a global static class named Environment. Below is a comparison between our proposed SFC and FC.}
	\vspace{-1.5em}
	\label{fig:VehicleWorld}
\end{figure*}

\subsection{Device}
\label{subsec:device}

As shown in Figure~\ref{fig:VehicleWorld}, we selected 30 common devices from intelligent cockpit systems, spanning four domains: Multimedia, Touch Control, Car Control, and Lighting. For each device, we collected commonly used APIs based on real-world usage (see Appendix~\ref{app:api_document} for API examples).

We abstracted each device into a corresponding module class by extracting relevant device properties for each API. These properties formed the foundation for constructing comprehensive object-oriented device classes. Within each class, we defined precise function signatures and implemented them with robust parameter validation and structured return formats. The API implementations leveraged a flexible combination of \texttt{get} and \texttt{set} methods operating on predefined attributes, effectively modeling the intended device behavior. This systematic approach resulted in the creation of 30 device classes encompassing 250 API methods and 680 attributes (Appendix~\ref{app:class} demonstrates detailed class definitions).

To support direct function calls, we also implemented two utility APIs inspired by \citet{trivedi2024appworld}: \texttt{search\_module} and \texttt{search\_api}. Agents use these to discover available modules and their APIs before executing functional calls.

\subsection{Global Attributes}
\label{subsec:global_attributes}

In contrast to the isolated smartphone applications in AppWorld~\citep{trivedi2024appworld}, intelligent cockpit systems operate as an ecosystem of tightly coupled devices that share and compete for limited system resources. This interdependence creates inherent challenges: the audio channel, for instance, can only be utilized by a single device at any given time, leading to potential conflicts among music playback, navigation instructions, and radio broadcasts. Similarly, system-wide properties such as volume represent global parameters that can be modified through multiple APIs, necessitating careful coordination to maintain system consistency and appropriate user experience.

To manage shared resources, we implemented a global \texttt{Environment} class using the Singleton pattern. This centralized component maintains system-wide attributes including sound channel, volume, and cabin temperature. It provides standardized access methods with concurrency control to prevent conflicts. Device classes must interact with shared attributes exclusively through this interface rather than maintaining local copies. Implementation details are provided in Appendix~\ref{app:environment}.

\subsection{World State}
\label{subsec:world_state}

To simulate diverse cockpit conditions, we implemented \texttt{init()} methods across all 30 device classes based on real-world usage patterns. These methods establish initial values for device attributes, creating consistent and realistic starting states. Through strategic combinations of these initialization methods, we generated 302 diverse initialization scenarios.
Furthermore, we serialized the relevant device states for each scenario into structured JSON documents. These documents capture both device-level attributes (e.g., navigation routes, music playback) and global settings (e.g., volume, temperature, sound channel). This standardized format enables precise scenario reproduction and supports systematic evaluation of state-based function call. In Table~\ref{table:state-document}, we demonstrate two example JSON documents.

\section{State-based Function Call}
\label{sec:sbfc}

Through the precise modeling of system states in VehicleWorld, we are able to analyze how the system evolves before and after function calls. We observe that in many scenarios, tasks can be accomplished by generating short state transitions instead of full API call sequences. Based on this insight, we propose State-based Function Call (SFC), which predicts the target system state and generates minimal code to fulfill the user's intent.

\subsection{Definition of Function Call}
\label{sec:definition_fc}

In the Function Call (FC) paradigm, the agent completes tasks by generating and executing sequences of API calls. Each function call \( f_i \) is formally defined as \( f_i = \mathcal{F}(q_i, A) \), where \( q_i \) represents the user query at step \( i \), and \( A \) denotes the set of available APIs. The agent interprets the query, selects a relevant API from \( A \), and generates a structured function call with the appropriate parameters. In our setup, the agent first invokes \texttt{search\_module} and \texttt{search\_api} to retrieve the list of available devices and their corresponding APIs (see Figure~\ref{fig:VehicleWorld}). Based on the retrieved APIs, it then constructs and executes function calls to fulfill the user's intent.

\subsection{Definition of State-based Function Call}
\label{sec:definition_sbfc}

In the State-based Function Call (SFC) paradigm, the agent completes tasks by directly predicting desired system states and generating efficient transition code. At each step \( i \), the agent processes the user query \( q_i \) alongside the current system state \( s_i \) to predict the subsequent state:

\begin{equation}
s_{i+1} = \mathcal{SF}(q_i, s_i),
\end{equation}
where \( \mathcal{SF} \) represents the state transition function, \( s_i \) is provided in the JSON format described in Section~\ref{subsec:world_state}, and \( s_{i+1} \) is the target state. This approach enables the agent to generate concise transition code that precisely fulfills user intent. In our implementation, we adopt a two-stage approach. First, we provide the agent with the complete JSON state information of all devices, enabling it to select the relevant devices that need to be operated based on the current state and user query. After identifying the pertinent devices, we then provide the agent with the specific state JSON of only those selected devices to predict the target state and generate the corresponding state transition code. Note that these two stages operate independently without shared context.



\section{Vehicle Benchmark}
\label{sec:vehicle_benchmark}

To evaluate model performance in VehicleWorld, we developed a comprehensive benchmark comprising natural, diverse, and challenging intelligent cockpit scenarios. Our approach includes a specialized data generation pipeline that maintains state continuity across interactions and an evaluation methodology that analyzes state transition patterns. 

\begin{figure*}[t]
	\raggedleft
	\centering
	\includegraphics[width=1.0\textwidth]{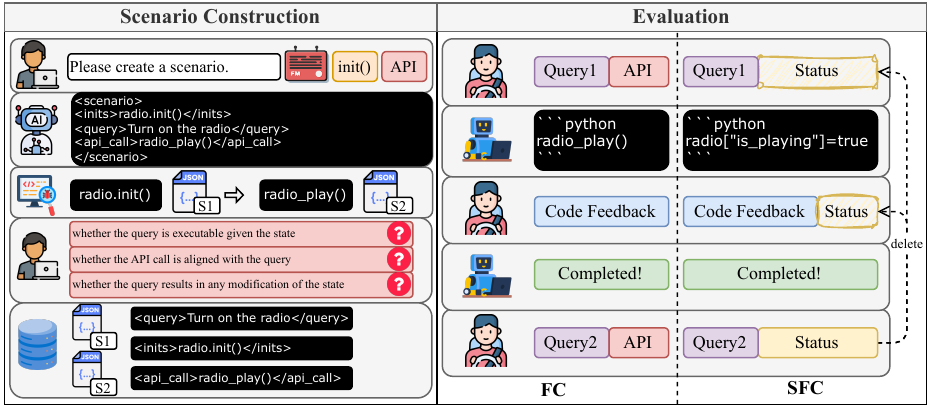}
	\caption{\textbf{Overview of Scenario Construction and Evaluation.} 
The left part shows the process of scenario generation, including device initialization, query creation, API selection, and validation through execution and expert review. The right part illustrates the evaluation pipeline, where FC executes API calls directly, while SFC manipulates system states explicitly and incorporates state information into multi-turn interactions.}
	\vspace{-1em}
	\label{fig:benchmark}
\end{figure*}
\begin{table*}[t]
	\centering
	\small
	\begin{tabular}{p{4.9cm}p{4.9cm}p{4.9cm}}
\cmidrule[\heavyrulewidth]{1-3}
\multicolumn{3}{p{14.7cm}}{
\textbf{Scenario 1:} Navigating to Shanghai via Nanjing, with the navigation volume set to 80 and no video playback.
} \\
\midrule
\fixedimg{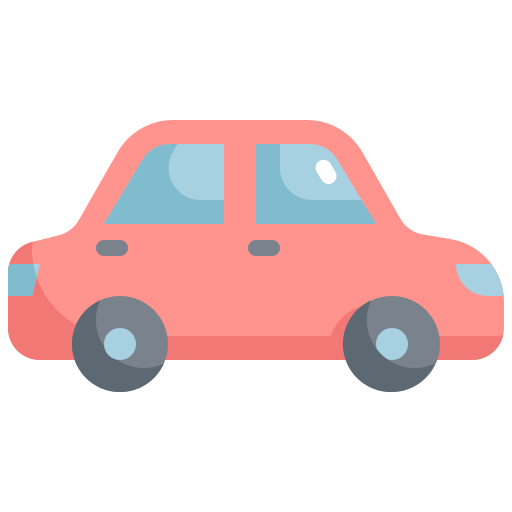}\xspace\textbf{Environment} \newline
$\bullet$ "volume": 80 \newline
$\bullet$ "sound\_channel": "navigation"
&
\fixedimg{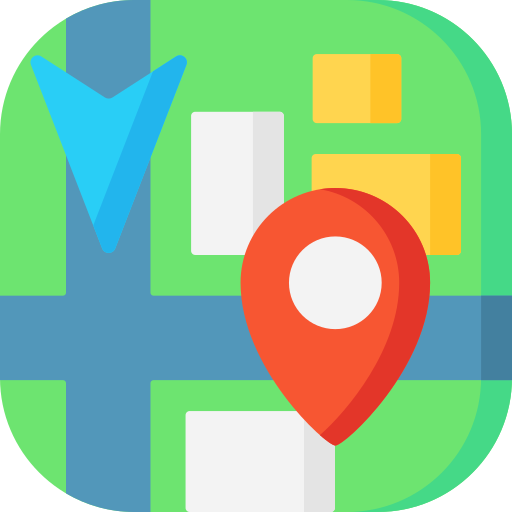}\xspace\textbf{Navigation} \newline
$\bullet$ "is\_active": true \newline
$\bullet$ "destination": "Shanghai" \newline
$\bullet$ "midway": "Nanjing"
&
\fixedimg{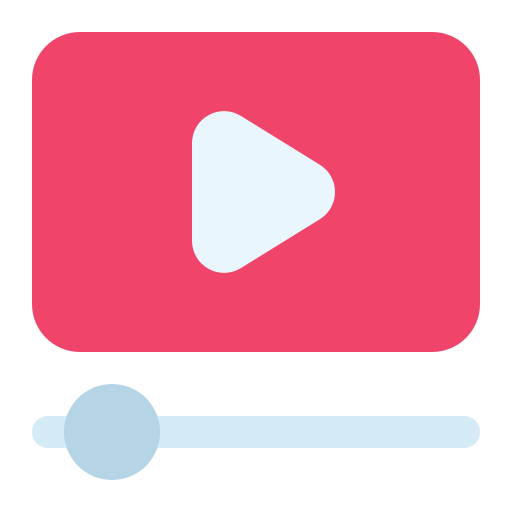}\xspace\textbf{Video} \newline
$\bullet$ "is\_playing": false \newline
$\bullet$ "quality": "1080P" \newline
$\bullet$ "current\_video": null
\\
\cmidrule[\heavyrulewidth]{1-3}
\multicolumn{3}{p{14.7cm}}{
\textbf{Scenario 2:} Turn on the air conditioner and lower it to 20 degrees, close the car door.
} \\
\midrule
\fixedimg{images/app-icons/car}\xspace\textbf{Environment} \newline
$\bullet$ "temperature": 20 \newline
$\bullet$ "speaker":"driver's seat"
&
\fixedimg{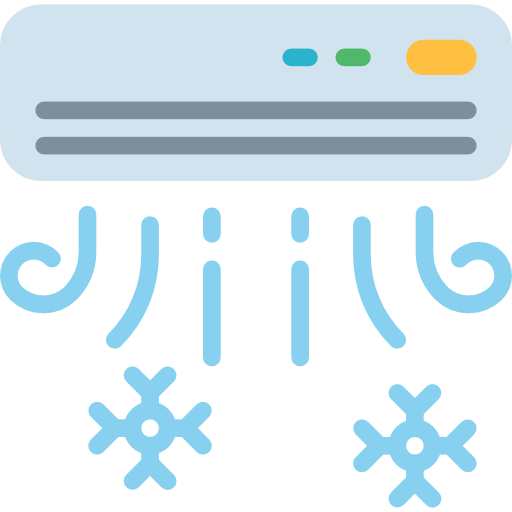}\xspace\textbf{AirConditioner} \newline
$\bullet$ "is\_on": true \newline
$\bullet$ "temperature": 20
&
\fixedimg{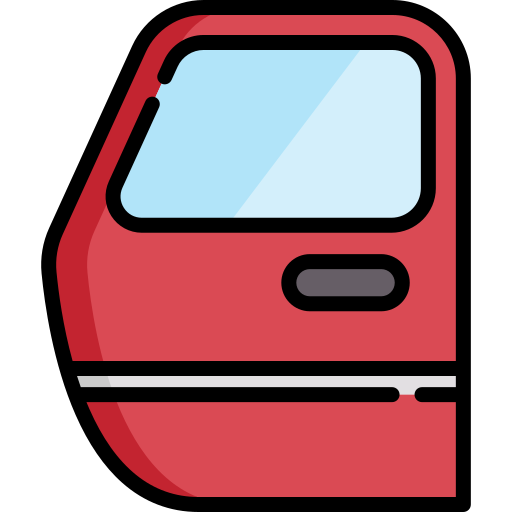}\xspace\textbf{Door} \newline
$\bullet$ "is\_locked": false \newline
$\bullet$ "status": "closed"
\\
    \bottomrule
	\end{tabular}
	\caption{
\textbf{State document}. Two common user scenarios in intelligent driving systems, each corresponding to different device configurations and operational states. These scenarios are stored as JSON files within the system.
}
	\label{table:state-document}
    \vspace{-2em}
\end{table*}

\subsection{Setup}
\label{subsec:setup}
To construct realistic and diverse scenarios, we begin by selecting appropriate devices and their associated APIs based on real-world user cases. For each scenario, we select the relevant devices and corresponding \texttt{init()} method and API. Then, we prompt Claude 3.7 Sonnet to generate a structured scenario following our predefined format (see Appendix~\ref{app:prompts}), as illustrated in Figure~\ref{fig:benchmark} (left).

As illustrated in Figure~\ref{fig:benchmark}, the generated scenario is organized using HTML-style tags: \texttt{<scenario>} wraps the entire scenario, \texttt{<inits>} defines the initialization methods, \texttt{<query>} specifies the user request, and \texttt{<api\_call>} represents the API call needed to fulfill the query. Since VehicleWorld is an executable environment, we execute both the \texttt{<inits>} and \texttt{<api\_call>} blocks to validate code correctness. Each execution step modifies the state of the relevant devices, and we persist the resulting intermediate states for subsequent evaluation.

In addition to automatic execution checks, each scenario undergoes manual inspection by domain experts. They verify (1) whether the query is executable given the current state, (2) whether the API call is semantically aligned with the query, and (3) whether the query results in any meaningful modification of the system state. Only scenarios that pass all validation criteria are stored in our benchmark database, including both the original structured scenario and all execution state records.

\subsection{VehicleWorld Benchmark}
\label{subsec:final dataset}

Through the aforementioned construction process and expert review, our final dataset contains 1291 tasks, split into Multimedia, 
Touch Control, Car Control, and Light splits, based on the types of devices they involve. Table~\ref{table:data-analysis} presents the distribution of user intents sampled from real-world scenarios, highlighting a significant diversity in the types of interactions, reflecting the wide range of devices involved. Car Control tasks dominate our dataset due to the large number of vehicle control systems and the critical nature of driving and safety functions, which demand frequent and varied interactions. The data also shows a strong preference for multi-round interactions, as users often refine commands or adjust settings through follow-ups.

The key dataset statistics are presented in Table~\ref{table:data-statistics}. On average, each task engages more than two devices, utilizes at least four distinct API calls, and involves approximately 3.5 API calls in total. The most complex scenarios within our dataset orchestrate up to five devices and 13 API calls, 12 of which are unique. This highlights the complexity of our dataset, underscoring its challenging nature.

\begin{table}[t]
    \centering
    \small
    \begin{tabular}{lcccc}\toprule
        & S-S & S-M & M-S & M-M \\
        \midrule
        Multimedia &59 & 96& 251& 53\\
        Touch Control &34& 93& 107& 15 \\
        Car Control &135& 178& 205& 173 \\
        Light & 79& 75& 159& 161\\
        \bottomrule
    \end{tabular}
    \caption{
      Distribution of categories in VehicleWorld Benchmark. The first S/M indicates single-turn/multi-turn interactions, while the second S/M indicates single/multiple intents per turn.
    }
    \label{table:data-analysis}
\end{table}
\begin{table}[t]
    \centering
    \small
    \begin{tabular}{lcccc}
        \toprule
        & \multicolumn{1}{c}{TC} & \multicolumn{1}{c}{L} & \multicolumn{1}{c}{M} & \multicolumn{1}{c}{CC}\\ 
        \midrule
        Avg Devices & 2.03 & 2.02 & 2.11 & 2.06 \\
        Avg Unique APIs & 2.84 & 3.54 & 3.18 & 3.47 \\
        Avg API Calls & 3.07 & 3.86 & 3.34 & 3.73 \\
        \bottomrule
    \end{tabular}
    \caption{Statistics of VehicleWorld Benchmark across devices. TC refers to Touch Control, L refers to Light, M refers to Multimedia and CC refers to Car Control.}
    \label{table:data-statistics}
    \vspace{-1em}
\end{table}

\subsection{Metrics}
\label{benchmark:metrics}



In evaluation, the key insight is that a good assistant should execute user-required actions accurately while refraining from undesired actions. Based on this principle, we introduce three evaluation criteria: (1) Whether attributes that should be changed have indeed been modified; (2) Whether attributes that should remain same are maintained consistently; and (3) Whether the trends of attribute changes are correct. Initially, we determine the expected attribute change by analyzing the differences between two consecutive interaction rounds of truth states, thus identifying sets of attributes that should and should not change. Subsequently, we compute the model-induced attribute change trends by analyzing the resulting states from interactions.

We propose three critical metrics: \textbf{F1 positive}, evaluates the model's effectiveness in accurately identifying attributes that require changes; \textbf{F1 negative}, measures the model’s capability in preserving attributes that should remain unchanged; and \textbf{Accuracy (Acc)}, indicates the proportion of attributes with correct change among all attributes intended for modification. The detailed computation formulas for these metrics are provided in Appendix~\ref{app:metric-detail}.

During evaluation, we prompt models to sequentially respond to user queries through multi-turn interactions, generating either API calls (FC) or state transitions (SFC). The agent's code responses are executed in a local environment, with return values and logs provided as feedback. In SFC, current device states are appended to the feedback for richer context. The final evaluation scores are computed by averaging results across all interaction turns. More evaluation details in Appendix~\ref{app:evaluation-detail}.

\section{Experiments}
\label{sec:experiments}
\begin{figure}[t]
    \raggedright
    \includegraphics[width=0.5\textwidth]{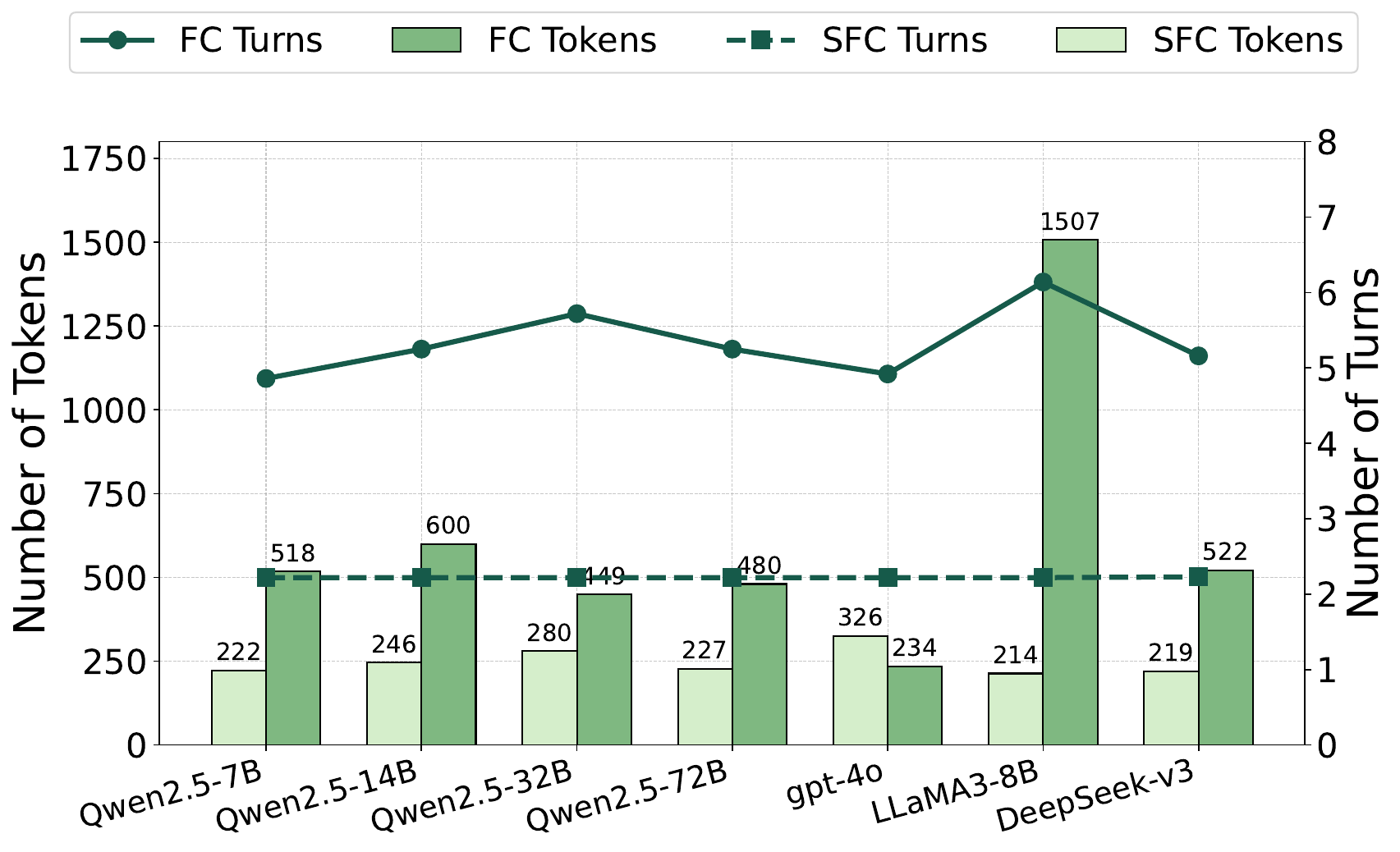}
    \caption{Average interaction turns and output tokens per task for each model under FC and SFC.}
    \label{fig:efficiency}
    \vspace{-.5em}
\end{figure}

\subsection{Experimental Setup}
\paragraph{Methods.} 
We adopted three prompting-based decision-making strategies in our experiments under both the Function Call (FC) and State-based Function Call (SFC) paradigms: ReAct, ReAct + Reflection, and ReAct without Examples. Detailed descriptions are provided in Appendix~\ref{app:methods}.

\vspace{-.5em}
\paragraph{Models.} We analyzed a range of open-source and commercial models, detailed in Appendix~\ref{app:model-detail}.

\subsection{Results}
\label{subsec:main-results}
\begin{table*}[t]
\belowrulesep=0pt
\aboverulesep=0pt
\fontsize{14}{21}\selectfont
\centering

\resizebox{\textwidth}{!}{
\begin{tabular}{l c|cccc| cccc| cccc| cccc}
\toprule[1.5pt]
\multirow{2}{*}{\textbf{Model Name}}& \multirow{2}{*}{\textbf{Overall}}  & \multicolumn{4}{c|}{\textbf{Multimedia}}  & \multicolumn{4}{c|}{\textbf{Touch Control}}& \multicolumn{4}{c|}{\textbf{Car Control}}& \multicolumn{4}{c}{\textbf{Light}}\\
\cmidrule(lr){3-6} \cmidrule(lr){7-10} \cmidrule(lr){11-14} \cmidrule(lr){15-18}
&& \textbf{S-S} & \textbf{S-M} & \textbf{M-S} & \textbf{M-M} &\textbf{S-S} &\textbf{S-M} &\textbf{M-S} &\textbf{M-M} & \textbf{S-S} & \textbf{S-M} & \textbf{M-S} & \textbf{M-M} & \textbf{S-S} & \textbf{S-M} & \textbf{M-S} & \textbf{M-M}  \\
 \midrule

\hline \multicolumn{18}{c}{\textit{\textbf{Function Call}}} \\   \hline 
Claude-3.7-Sonnet & 62.9 & 74.3 & 40.0 & 66.5 & 48.5 & 50.5 & 31.4 & 90.3 & 50.6 & 65.0 & 38.7 & 51.9 & 55.0 & 57.1 & 36.0 & 79.0 & 59.5 \\
GPT-4o & \textbf{70.2} & \textbf{77.1} & \textbf{66.1} & 68.4 & 56.4 & \textbf{78.1} & 45.6 & 92.4 & 58.4 & \textbf{69.9} & \textbf{60.5} & 55.9 & 56.4 & \textbf{75.3} & \textbf{60.3} & 81.6 & 62.2 \\
DeepSeek-v3-250324 & 70.0 & 65.2 & 58.1 & \textbf{70.1} & \textbf{59.0} & 67.4 & \textbf{48.0} & \textbf{92.9} & \textbf{60.6} & 68.4 & 56.2 & \textbf{58.6} & \textbf{62.2} & 73.4 & 56.3 & \textbf{82.1} & \textbf{66.9} \\
Qwen2.5-7B & 33.0 & 65.0 & 32.4 & 49.7 & 32.3 & 32.3 & 11.9 & 66.6 & 15.8 & 17.3 & 10.9 & 28.2 & 21.9 & 9.1 & 8.8 & 29.6 & 23.1 \\
Qwen2.5-14B & 38.0 & 59.3 & 33.3 & 52.5 & 33.8 & 32.3 & 13.0 & 68.1 & 24.1 & 33.8 & 13.1 & 35.6 & 29.2 & 11.7 & 9.7 & 36.1 & 29.3 \\
Qwen2.5-32B & 46.8 & 62.1 & 39.6 & 58.9 & 35.6 & 40.6 & 14.6 & 77.0 & 24.8 & 44.4 & 22.9 & 42.1 & 41.9 & 22.1 & 11.8 & 53.1 & 44.7 \\
Qwen2.5-72B & 58.8 & 69.9 & 41.7 & 63.1 & 46.5 & 46.9 & 15.6 & 85.6 & 46.7 & 60.2 & 29.6 & 50.8 & 56.3 & 50.6 & 26.4 & 72.9 & 58.3 \\
Llama-3.1-8B & 28.7 & 50.0 & 22.1 & 42.5 & 31.0 & 26.0 & 12.5 & 59.5 & 15.8 & 13.5 & 9.7 & 25.8 & 20.1 & 10.4 & 8.1 & 26.7 & 21.6 \\
\cellcolor{gray!10} \textbf{Avg.} & \cellcolor{gray!10} 51.1 & \cellcolor{gray!10} 65.4 & \cellcolor{gray!10} 41.7 & \cellcolor{gray!10} 59.0 & \cellcolor{gray!10} 42.9 & \cellcolor{gray!10} 46.8 & \cellcolor{gray!10} 24.1 & \cellcolor{gray!10} 79.1 & \cellcolor{gray!10} 37.1 & \cellcolor{gray!10} 46.6 & \cellcolor{gray!10} 30.2 & \cellcolor{gray!10} 43.6 & \cellcolor{gray!10} 42.9 & \cellcolor{gray!10} 38.7 & \cellcolor{gray!10} 27.2 & \cellcolor{gray!10} 57.6 & \cellcolor{gray!10} 45.7 \\

\hline \multicolumn{18}{c}{\textit{\textbf{State-based Function Call}}} \\   \hline 

Claude-3.7-Sonnet & \textbf{73.4} & \textbf{72.3} & \textbf{66.6} & \textbf{74.0} & 55.6 & 73.8 & \textbf{46.4} & \textbf{94.1} & \textbf{72.8} & 69.9 & \textbf{68.2} & \textbf{63.2} & \textbf{65.7} & 74.7 & 58.0 & \textbf{83.2} & \textbf{69.8} \\
GPT-4o & 70.9 & 67.2 & 57.3 & 72.0 & \textbf{56.1} & 69.8 & 36.3 & 90.9 & \textbf{72.8} & 69.3 & 55.1 & 57.3 & 61.3 & \textbf{77.3} & 51.1 & 79.7 & 64.2 \\
DeepSeek-v3-250324 & 71.9 & 67.1 & 63.9 & 71.6 & 54.3 & 74.9 & 43.6 & 93.7 & 69.4 & \textbf{72.3} & 60.3 & 59.7 & 61.9 & 74.7 & \textbf{60.0} & 81.6 & 66.0 \\
Qwen2.5-7B & 49.0 & 60.3 & 32.9 & 55.4 & 34.9 & 60.9 & 12.5 & 73.9 & 18.9 & 46.0 & 18.4 & 33.4 & 32.1 & 61.4 & 23.7 & 65.4 & 37.3 \\
Qwen2.5-14B & 57.3 & 66.8 & 33.7 & 59.8 & 45.0 & 62.0 & 22.4 & 82.1 & 41.8 & 56.1 & 30.7 & 42.6 & 44.9 & 73.4 & 27.3 & 71.8 & 49.9 \\
Qwen2.5-32B & 57.5 & 65.4 & 42.1 & 65.1 & 47.3 & 55.7 & 18.4 & 81.8 & 36.5 & 36.1 & 27.3 & 45.7 & 50.6 & 72.7 & 36.4 & 76.1 & 56.2 \\
Qwen2.5-72B & 64.8 & 67.5 & 47.2 & 65.4 & 49.7 & \textbf{76.6} & 34.8 & 91.1 & 55.1 & 58.4 & 47.0 & 53.2 & 52.3 & 76.6 & 44.4 & 78.2 & 55.9 \\
Llama-3.1-8B & 46.8 & 63.2 & 43.2 & 54.9 & 36.2 & 50.0 & 9.9 & 74.9 & 24.5 & 30.8 & 19.2 & 31.5 & 29.8 & 50.6 & 18.6 & 63.7 & 35.5 \\
\cellcolor{gray!10} \textbf{Avg.} & \cellcolor{gray!10} 61.5 & \cellcolor{gray!10} 66.2 & \cellcolor{gray!10} 48.4 & \cellcolor{gray!10} 64.8 & \cellcolor{gray!10} 47.4 & \cellcolor{gray!10} 65.5 & \cellcolor{gray!10} 28.0 & \cellcolor{gray!10} 85.3 & \cellcolor{gray!10} 49.0 & \cellcolor{gray!10} 54.9 & \cellcolor{gray!10} 40.8 & \cellcolor{gray!10} 48.3 & \cellcolor{gray!10} 49.8 & \cellcolor{gray!10} 70.2 & \cellcolor{gray!10} 39.9 & \cellcolor{gray!10} 75.0 & \cellcolor{gray!10} 54.3 \\
\bottomrule[1.5pt]

\end{tabular}}
\caption{
Accuracy comparison of models across different control domains and interaction complexities (S-S, S-M, M-S, M-M) under FC and SFC paradigms. Overall, SFC consistently outperforms FC across all domains and task.
}
\label{table:main-results}
\vspace{-2em}
\end{table*}

\cref{table:main-results} presents results across all evaluated models in four domains (Multimedia, Touch Control, Car Control, and Light), categorized by interaction complexity (S-S, S-M, M-S, and M-M) under both FC and SFC paradigms for the ReAct strategy. Models consistently demonstrate superior performance under SFC, with average accuracy increasing from 51.1\% (FC) to 61.5\% (SFC), underscoring the advantage of direct state prediction over sequential API manipulation. GPT-4o emerges as the top performer under the FC paradigm (70.2\%), while Claude-3.7-Sonnet performs best under the SFC paradigm (73.4\%). Domain-specific analysis reveals the most pronounced improvements in Touch Control and Car Control tasks. In Figure~\ref{fig:efficiency}, we further analyze the latency differences between FC and SFC from the perspectives of interaction rounds and generated token counts. We observe that SFC significantly reduces both the number of interaction rounds and generated tokens across various models, effectively enhancing execution efficiency and establishing a powerful baseline in vehicle-based LLM applications.

\begin{figure}[t]
    \raggedright
    \includegraphics[width=0.5\textwidth]{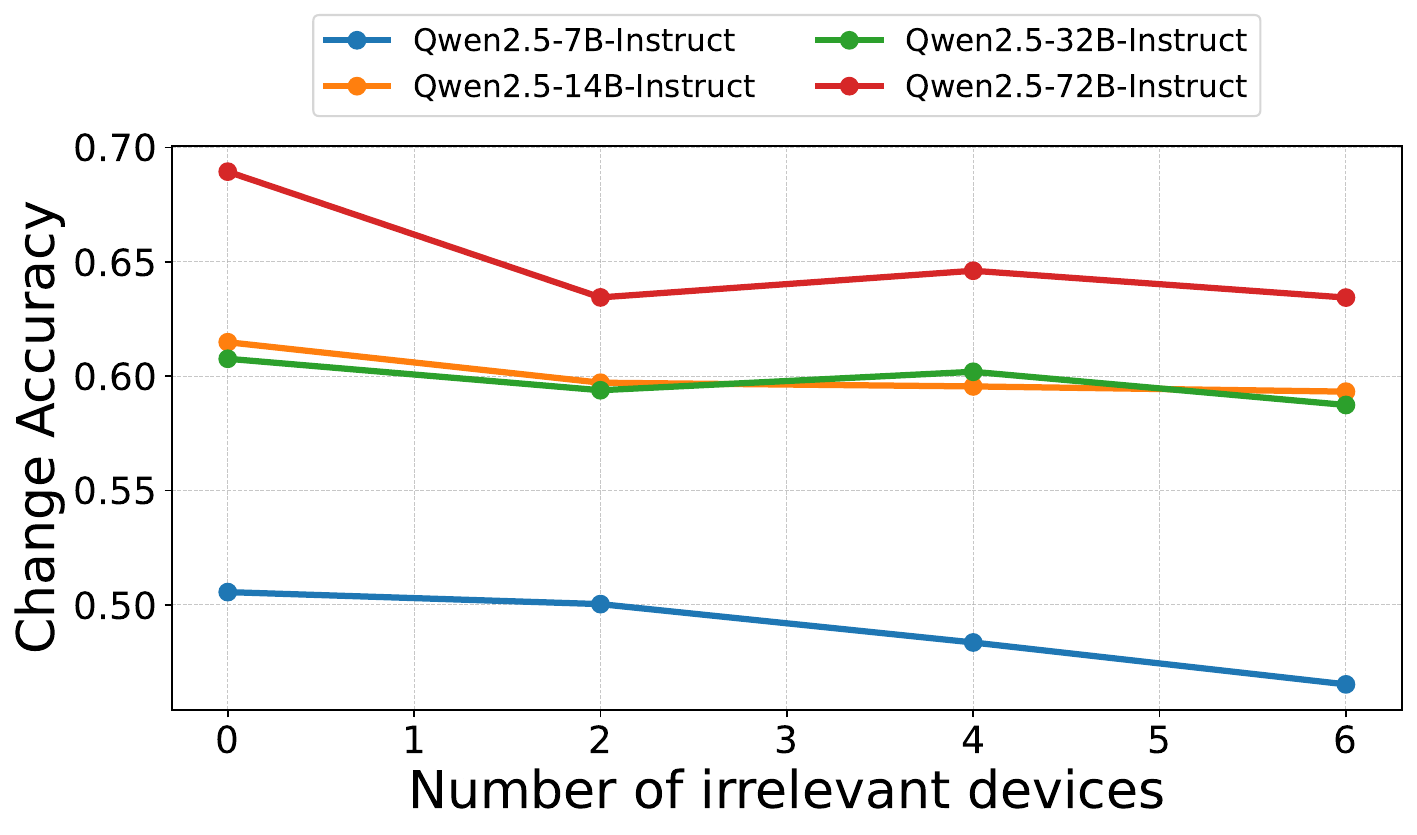}
    \caption{\textbf{Impact of World Complexity.} Accuracy across varying levels of world complexity, measured by the number of relevant devices.}
    \vspace{-.5em}
    \label{fig:Complexity}
\end{figure}

\subsection{Analysis}
\label{subsec:analysis}
\paragraph{World Complexity.} Figure~\ref{fig:Complexity} illustrates Qwen2.5 series performance across increasing world complexity. We simulate complexity gradients by incrementally adding 2, 4, or 6 irrelevant device states to the ReAct prompting setup, resulting in consistent performance degradation across all models. Notably, larger models demonstrate superior robustness as world complexity increases, while the Qwen2.5-7B model exhibits rapid performance deterioration. This pattern suggests that increased parameter scale enhances models' capacity to comprehend and navigate complex environments.

\begin{figure}[t]
    \raggedright
    \includegraphics[width=0.5\textwidth]{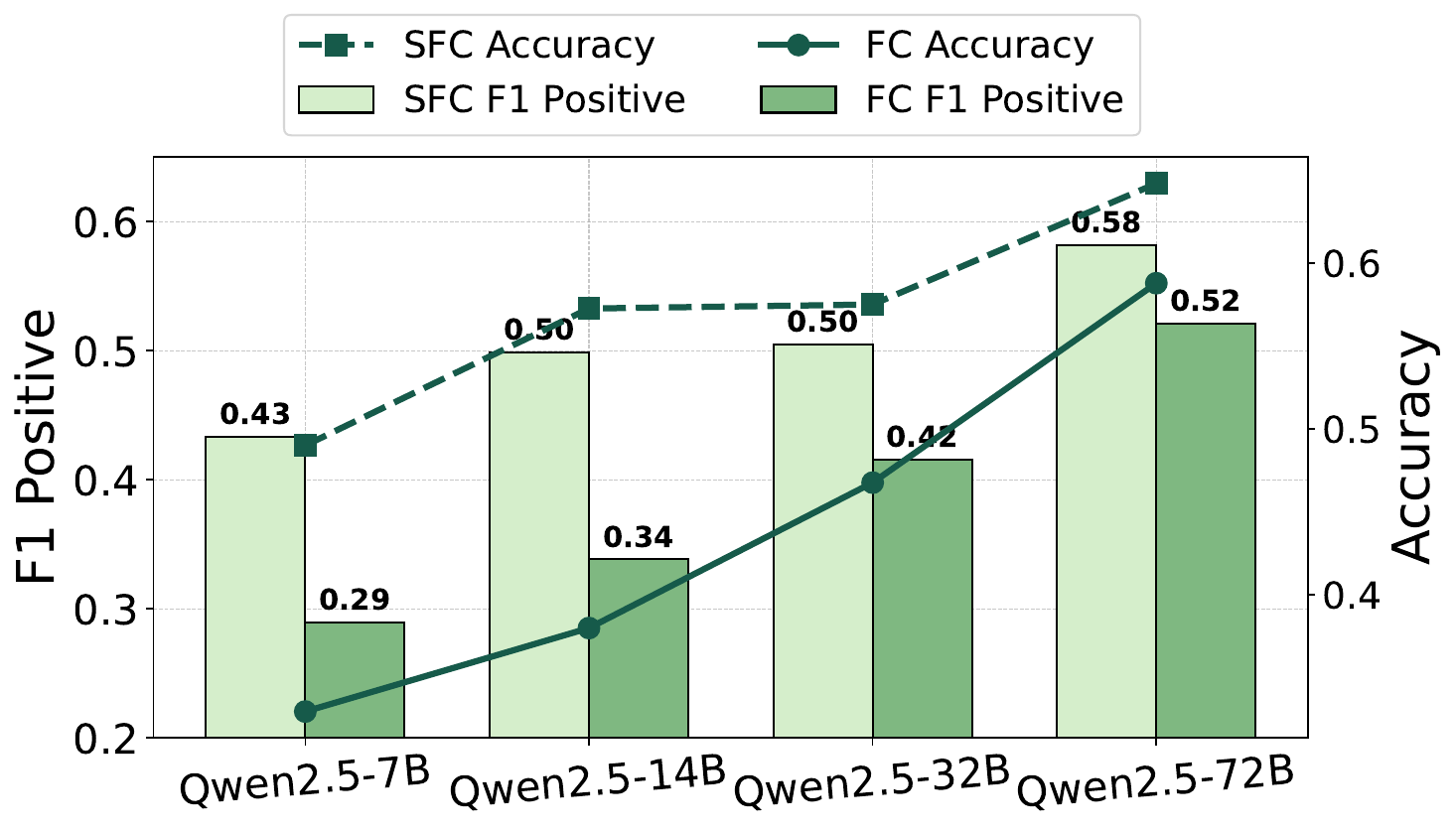}
    \caption{\textbf{Scaling Effects in VehicleWorld.} Performance comparison (Accuracy and F1 Score) across different model parameter scales.}
    \vspace{-.5em}
    \label{fig:Scaling}
\end{figure}

\begin{figure*}[t]
	\raggedleft
	\centering
	\includegraphics[width=1.0\textwidth]{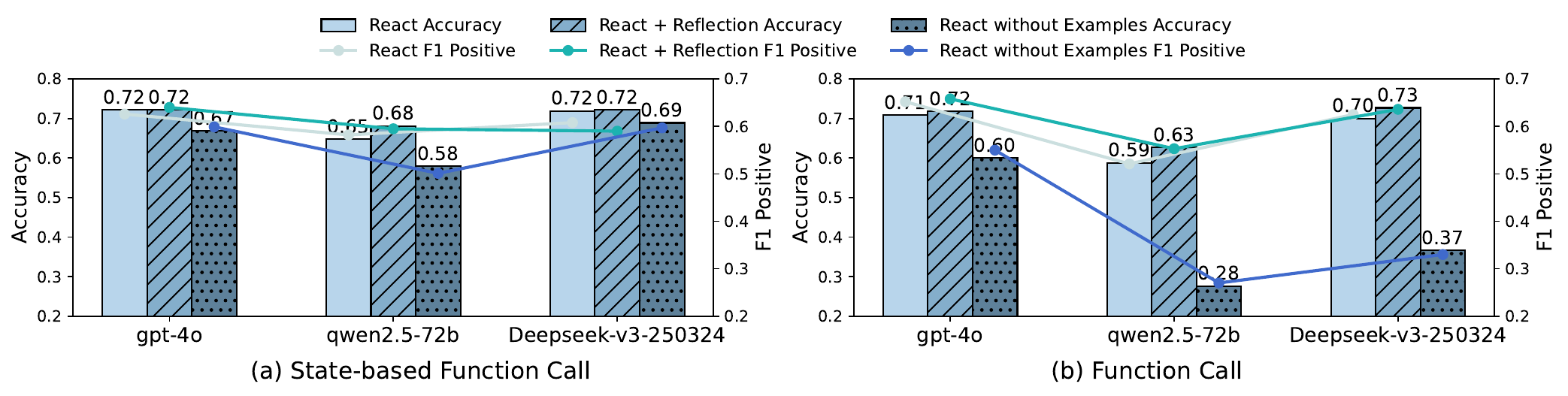}
	\caption{\textbf{Comparison of prompting strategies under the SFC and FC paradigms.} Each bar group corresponds to a model evaluated with ReAct, ReAct + Reflection, and ReAct without Examples. Accuracy is shown as bars (left axis), and F1 score on positive classes is shown as lines (right axis).}
	\vspace{-2em}
	\label{fig:method}
\end{figure*}

\vspace{-.5em}
\paragraph{Scaling Effects.} Figure~\ref{fig:Scaling} demonstrates the impact of model scaling on both SFC and FC performance. As Qwen2.5 model size increases from 7B to 72B parameters, both methods show improved performance. FC exhibits larger relative gains with scaling, while SFC consistently outperforms FC across all model scales. 

\vspace{-.5em}
\paragraph{Impact of Reflection and Examples.}
Figure~\ref{fig:method} compares three prompting strategies: ReAct, ReAct with Reflection, and ReAct without Examples, evaluated under both the SFC and FC paradigms for representative open-source and commercial models. We observe that removing in-context examples degrades performance in both paradigms, but the impact is significantly greater in FC. This suggests that SFC is more robust to the absence of demonstrations. Additionally, adding reflection consistently improves results in both SFC and FC. More experimental results can be found in Table~\ref{tables:reflect-results} and~\ref{tables:nosample-results}.

\begin{figure}[t]
    \raggedright
    \includegraphics[width=0.5\textwidth]{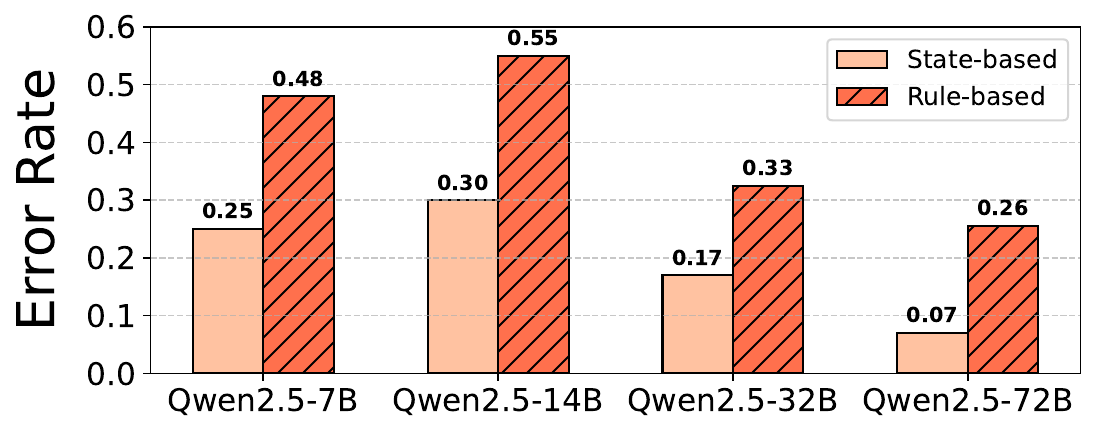}
    \caption{Error rate comparision between state-based and rule-based evaluation.}
    \label{fig:error-rate}
    \vspace{-.5em}
\end{figure}

\vspace{-.5em}
\paragraph{State-based vs Rule-based Evaluation.} 
A major advantage of constructing executable environments is the ability to accurately evaluate agent behaviors. In Figure~\ref{fig:error-rate}, we compare the differences between state-based and traditional rule-based evaluations for the Qwen-2.5 model. Both automated evaluation methods are detailed in Appendix~\ref{app:error-rate}. By comparing with expert evaluation results, we find that state-based evaluation consistently results in lower error rates across all model sizes. The lower error rate under state-based evaluation can be attributed to its emphasis on task completion rather than strict adherence to predefined action sequences, providing a more nuanced assessment of agent performance in interactive environments.

\begin{figure}[t]
    \raggedright
    \includegraphics[width=0.5\textwidth]{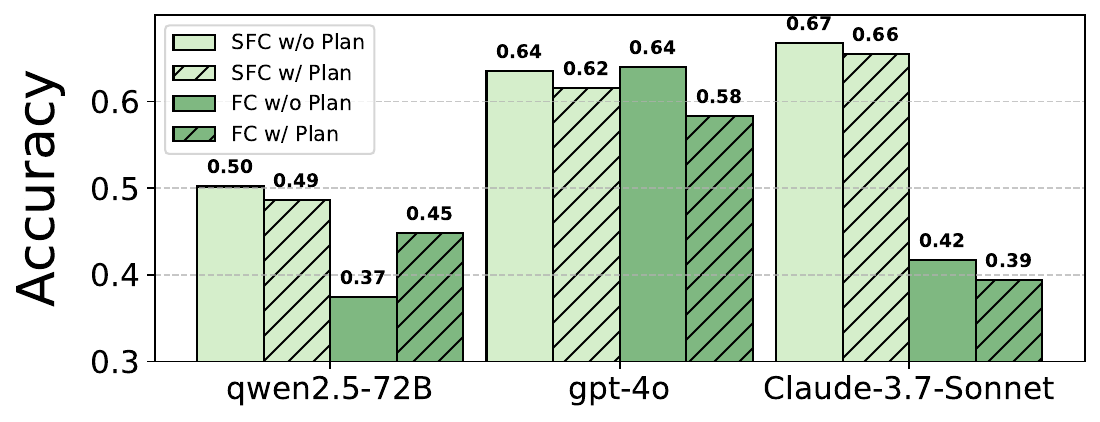}
    \caption{Performance of various models under the FC and SFC paradigms, with and without reasoning.}
    \label{fig:reasoning}
    \vspace{-.5em}
\end{figure}

\vspace{-.5em}
\paragraph{Reasoning.}  
In Figure~\ref{fig:reasoning}, we analyze the impact of extended thinking on task accuracy. We implemented ReAct prompting strategies to encourage more comprehensive planning and reasoning processes. The results reveal nuanced patterns: under the SFC paradigm, all models experience performance degradation when reasoning is introduced, suggesting that the streamlined execution may be disrupted by excessive deliberation. However, under the FC paradigm, we observe contrasting effects—while GPT-4o and Claude-3.7-Sonnet show performance decline, Qwen2.5-72B demonstrates improvement with reasoning. This suggests that models with initially lower performance in complex interaction scenarios may benefit from structured planning, whereas high-performing models may suffer from overthinking~\citep{chen2024not}, whereby they generate fabricated facts that contradict the actual environment state. This phenomenon highlights the balance between sufficient reasoning and excessive deliberation in agent-based systems, with the optimal strategy varying by model capability and execution paradigm.

\begin{figure*}[t]
    \centering
    \includegraphics[width=\textwidth]{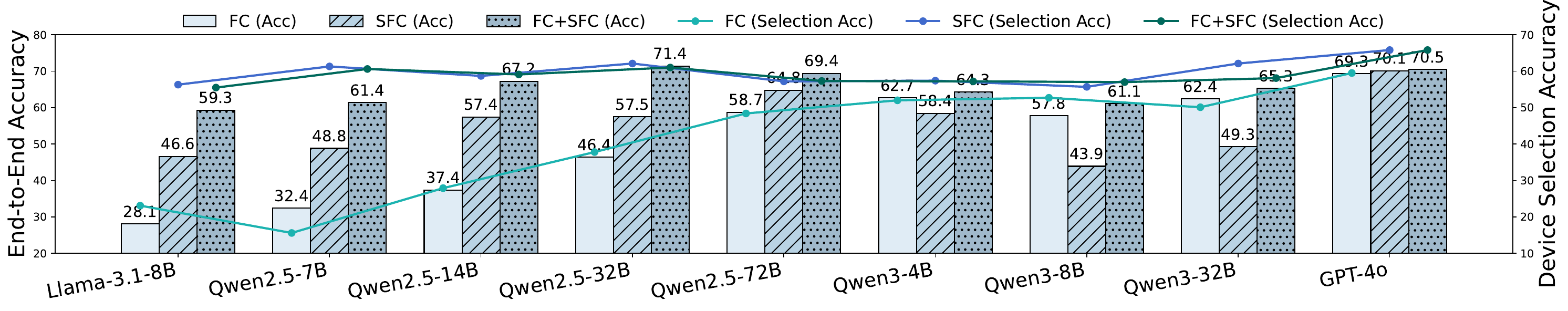} 
    \caption{\textbf{Comparison of FC, SFC, and the Integration of FC and SFC.} Each bar group shows end-to-end accuracy for FC, SFC, and FC+SFC (bars, left axis), with device selection accuracy shown as lines (right axis).}
    \vspace{-1.5em}
    \label{fig:FC+SFC}
\end{figure*}

\vspace{-.5em}
\paragraph{The Integration of FC and SFC.} 
To further investigate factors contributing to SFC's high accuracy, we conducted a detailed analysis comparing error cases between FC and SFC methods. Our findings reveal that SFC significantly outperforms FC in device selection accuracy. For ambiguous user queries, FC may require multiple rounds of environment exploration to determine target devices, whereas SFC, with its global environment perception, can more accurately identify target devices.

However, we observed scenarios where SFC underperforms compared to FC. These cases typically occur when numerous devices require extensive state transition code generation. In complex device states, FC benefits from high-level API encapsulation, enabling easy manipulation of multiple device properties and achieving higher accuracy.

Based on these observations, we analyze the integration of two approaches (FC+SFC): leveraging SFC's environment perception capability for device selection, then providing relevant device APIs for FC-based calls. The experimental results are presented in Figure~\ref{fig:FC+SFC}. Experimental results show that the FC+SFC method achieves the highest end-to-end accuracy. Notably, Qwen3 series models, due to specialized tool calling training, demonstrate that FC accuracy can exceed SFC accuracy, showcasing their powerful tool calling capabilities even with lower device selection accuracy.

\section{Conclusion}
\label{sec:conclusion}
We introduced VehicleWorld, the first comprehensive multi-device environment for intelligent vehicle interaction that accurately models the complex, interconnected systems in modern cockpits. 
This environment enables precise evaluation of agent behaviors by providing real-time state information during execution. 
Building on this foundation, we constructed the first benchmark for evaluating vehicle agents, establishing standardized metrics for comparing different LLMs. 
Our analysis revealed a critical insight: directly predicting environment states proves more effective than predicting function calls in complex, state-dependent systems. 
This observation led to our State-based Function Call (SFC) approach, which maintains explicit awareness of system state and implements direct state transitions.

Experimental results demonstrate that SFC significantly outperforms traditional function calling, enhancing models' ability to interpret user intentions while reducing erroneous function calls. Furthermore, we identified the complementary strengths of both paradigms: SFC excels at device selection due to its global environmental perception, while FC's high-level APIs are more efficient for complex state transitions. This led us to propose a hybrid FC+SFC approach, which leverages SFC for device selection before using FC for API execution. Our experiments confirm that this hybrid method achieves the highest end-to-end accuracy.

These advancements establish a foundation for future research in intelligent cockpit systems and offer valuable insights for agent design in other complex, multi-device environments.


\section*{Limitations}

\paragraph{Manual Entity Construction.} While our object-oriented approach to designing and implementing the world model provides a highly executable, persistent, and inspectable simulation environment, it still has limitations. Specifically, the lack of an efficient automated pipeline for constructing entity classes requires substantial manual effort to verify and define each device. Additionally, we have not yet developed an effective method for integrating the global static \texttt{Environment} class into the construction of individual entities. Future work will explore more automated and standardized pipelines for world model construction.

\paragraph{Complex State Representation.} Moreover, although we adopt the JSON format to store and present the world state, the complexity of the environment forces us to augment each attribute with a \texttt{value}, \texttt{type}, and \texttt{description} field to help the agent understand the meaning of each parameter. Future research could explore methods to streamline the world state representation or improve the agent's ability to interpret complex environments. As discussed in Section~\ref{subsec:analysis}, our experiments show that model performance tends to degrade as world complexity increases. Enhancing an agent's understanding of complex environments may therefore improve its performance in state-based function call (SFC) tasks under such conditions. 
\section*{Ethics Statement}

All APIs used in our system were designed and implemented by domain experts based on real-world user experiences in intelligent cockpit environments, combined with personal usage habits. These APIs do not involve any third-party proprietary or private data and are intended strictly for research purposes. Although the APIs are expert-defined rather than crowd-sourced, they are derived from realistic production environments, and thus offer valuable insights for academic study.

The scenarios included in our benchmark were created by the participants from Pacific Rim region. When recruiting participants, we carefully considered gender and racial balance, while strictly adhering to local wage standards by compensating participants at a rate of \$20 per hour, in line with the regional average. We thoroughly informed all participants about the nature of their work, as shown in Tables~\ref{tab:api-design-instruction}, \ref{tab:test-set-instruction}, and~\ref{tab:function-call-evaluation}, as well as how their data would be subsequently utilized, as detailed in Table~\ref{tab:annotation-purpose}. To further expand our dataset, we employed Claude-3.7-Sonnet~\citep{Anthropic2025Claude37Sonnet} to generate additional scenarios following our predefined format. In the development of VehicleWorld, we utilized Claude-3.7-Sonnet to assist us with code generation. All AI-generated content was carefully reviewed by human annotators to ensure that the data contained no personally identifiable information or offensive content.

Our environment successfully simulates an executable, programmable vehicle cockpit system. However, it remains a controlled research simulation that may differ significantly from real-world deployments. Caution should be exercised when considering the use of this environment as a staging ground for in-vehicle agents in production systems. Transitioning from a simulated to a real-world environment raises critical concerns regarding execution safety, system reliability, and user interaction risks, all of which must be rigorously addressed in future work before deployment.

Finally, we have strictly ensured compliance with all relevant terms of use in our deployment of large language models. For commercial models, we exclusively utilized official APIs and adhered rigorously to their terms of service. For open-source models, we carefully reviewed all licenses and ensured our usage complies with their requirements; details of these licenses can be found in Table~\ref{tables:model-detail}. Throughout our data collection and usage processes, we have meticulously ensured compliance with relevant legal and ethical standards, and have provided detailed information to all data contributors regarding the purpose and scope of our data usage.


\bibliography{main}

\appendix
\section{Experiment Detail} 
\label{app:experiment-detail}
\subsection{Experiment Setting} 
\label{app:setting}
During inference, we adopted a uniform sampling temperature of 0.7 and deployed all open-source models using vLLM~\citep{kwon2023efficient} on 8 interconnected NVIDIA A100 GPUs. For the Qwen2.5 series, we extended the context length to 128k tokens using YaRN, a technique for improving length extrapolation. For Qwen3, we also applied YaRN to extend the context length and evaluated the model without activating thinking mode. All model calls were issued through the standard OpenAI SDK interface to ensure consistency across different model providers. The results remained highly stable across multiple runs. To reduce computational overhead without compromising reliability, we therefore adopted single-sample evaluation for all experiments.

\subsection{Methods} 
\label{app:methods}
\begin{table*}[t]
\belowrulesep=0pt
\aboverulesep=0pt
\fontsize{14}{21}\selectfont
\centering

\resizebox{\textwidth}{!}{
\begin{tabular}{l c|cccc| cccc| cccc| cccc}
\toprule[1.5pt]
\multirow{2}{*}{\textbf{Model Name}}& \multirow{2}{*}{\textbf{Overall}}  & \multicolumn{4}{c|}{\textbf{Multimedia}}  & \multicolumn{4}{c|}{\textbf{Touch Control}}& \multicolumn{4}{c|}{\textbf{Car Control}}& \multicolumn{4}{c}{\textbf{Light}}\\
\cmidrule(lr){3-6} \cmidrule(lr){7-10} \cmidrule(lr){11-14} \cmidrule(lr){15-18}
&& \textbf{S-S} & \textbf{S-M} & \textbf{M-S} & \textbf{M-M} &\textbf{S-S} &\textbf{S-M} &\textbf{M-S} &\textbf{M-M} & \textbf{S-S} & \textbf{S-M} & \textbf{M-S} & \textbf{M-M} & \textbf{S-S} & \textbf{S-M} & \textbf{M-S} & \textbf{M-M}  \\
 \midrule

\hline \multicolumn{18}{c}{\textit{\textbf{Function Call}}} \\   \hline 

GPT-4o & 71.8 & 77.1 & \textbf{66.1} & \textbf{68.8} & 55.8 & 69.8 & 47.2 & 92.6 & 68.8 & 69.2 & \textbf{63.6} & 57.2 & 58.8 & \textbf{79.2} & 58.6 & 82.4 & 63.6 \\
DeepSeek-v3-250324 & \textbf{72.7} & 70.4 & 62.4 & 68.2 & \textbf{57.5} & \textbf{87.5} & \textbf{50.0} & \textbf{94.3} & \textbf{70.8} & \textbf{72.4} & 62.3 & \textbf{58.8} & \textbf{64.2} & 77.3 & \textbf{58.9} & \textbf{82.5} & \textbf{68.3} \\
Llama-3.1-8B & 31.2 & 57.1 & 25.6 & 45.2 & 28.9 & 29.2 & 10.0 & 64.9 & 17.4 & 17.3 & 11.8 & 26.8 & 21.9 & 7.8 & 8.1 & 29.9 & 24.0 \\
Qwen-2.5-72B & 62.7 & \textbf{77.5} & 48.7 & 66.9 & 46.5 & 48.4 & 18.2 & 90.5 & 44.7 & 58.8 & 31.4 & 54.3 & 58.9 & 69.5 & 31.1 & 76.1 & 62.1 \\
Qwen-2.5-32B & 54.6 & 62.0 & 49.2 & 63.4 & 48.6 & 56.2 & 19.8 & 83.8 & 40.6 & 51.9 & 28.8 & 48.5 & 49.1 & 40.9 & 26.5 & 60.8 & 51.1 \\
Qwen-2.5-14B & 43.6 & 59.9 & 36.9 & 54.0 & 36.8 & 38.5 & 15.1 & 74.2 & 38.4 & 36.1 & 18.3 & 37.8 & 40.1 & 20.8 & 17.3 & 48.9 & 40.1 \\
Qwen-2.5-7B & 34.8 & 61.8 & 39.4 & 50.7 & 32.0 & 32.3 & 12.0 & 65.5 & 15.8 & 18.5 & 12.5 & 29.2 & 23.5 & 10.4 & 11.0 & 36.2 & 26.6 \\
\cellcolor{gray!10} \textbf{Avg.} & \cellcolor{gray!10} 53.1 & \cellcolor{gray!10} 66.5 & \cellcolor{gray!10} 46.9 & \cellcolor{gray!10} 59.6 & \cellcolor{gray!10} 43.7 & \cellcolor{gray!10} 51.7 & \cellcolor{gray!10} 24.6 & \cellcolor{gray!10} 80.8 & \cellcolor{gray!10} 42.4 & \cellcolor{gray!10} 46.3 & \cellcolor{gray!10} 32.7 & \cellcolor{gray!10} 44.7 & \cellcolor{gray!10} 45.2 & \cellcolor{gray!10} 43.7 & \cellcolor{gray!10} 30.2 & \cellcolor{gray!10} 59.5 & \cellcolor{gray!10} 48.0 \\

\hline \multicolumn{18}{c}{\textit{\textbf{State-based Function Call}}} \\   \hline

GPT-4o & \textbf{72.2} & 75.7 & 61.2 & \textbf{72.9} & 57.0 & 69.8 & 41.7 & \textbf{94.0} & 68.5 & 68.3 & \textbf{63.1} & \textbf{61.3} & 62.0 & 75.3 & 58.2 & 81.0 & 66.6 \\
DeepSeek-v3-250324 & \textbf{72.2} & \textbf{77.3} & \textbf{61.5} & 71.0 & \textbf{57.7} & \textbf{81.7} & \textbf{45.4} & 93.1 & \textbf{74.1} & \textbf{69.9} & 58.5 & 60.5 & \textbf{64.2} & 75.3 & \textbf{59.8} & \textbf{82.5} & \textbf{68.2} \\
Llama-3.1-8B & 48.3 & 60.5 & 37.8 & 56.7 & 39.1 & 56.8 & 11.1 & 77.1 & 37.6 & 32.1 & 20.8 & 35.9 & 34.1 & 51.9 & 21.8 & 63.0 & 40.2 \\
Qwen-2.5-72B & 68.1 & 70.4 & 56.1 & 70.2 & 54.8 & 75.9 & 35.0 & 92.3 & 62.5 & 63.2 & 52.4 & 57.3 & 56.7 & 73.4 & 51.7 & 79.4 & 61.6 \\
Qwen-2.5-32B & 68.1 & 71.8 & 53.7 & 68.7 & 55.7 & 66.1 & 30.1 & 91.2 & 59.8 & 65.7 & 48.8 & 55.0 & 59.8 & \textbf{75.9} & 51.9 & 78.4 & 63.4 \\
Qwen-2.5-14B & 62.1 & 65.3 & 43.8 & 62.8 & 46.5 & 72.5 & 29.9 & 85.1 & 49.6 & 61.7 & 41.3 & 47.2 & 51.1 & 72.8 & 33.5 & 73.4 & 54.9 \\
Qwen-2.5-7B & 52.9 & 61.2 & 39.9 & 55.4 & 40.0 & 60.9 & 18.2 & 76.7 & 30.7 & 52.2 & 31.3 & 38.4 & 34.9 & 58.5 & 31.6 & 67.2 & 40.5 \\
\cellcolor{gray!10} \textbf{Avg.} & \cellcolor{gray!10} 63.4 & \cellcolor{gray!10} 68.9 & \cellcolor{gray!10} 50.6 & \cellcolor{gray!10} 65.4 & \cellcolor{gray!10} 50.1 & \cellcolor{gray!10} 69.1 & \cellcolor{gray!10} 30.2 & \cellcolor{gray!10} 87.1 & \cellcolor{gray!10} 54.7 & \cellcolor{gray!10} 59.0 & \cellcolor{gray!10} 45.2 & \cellcolor{gray!10} 50.8 & \cellcolor{gray!10} 51.8 & \cellcolor{gray!10} 69.0 & \cellcolor{gray!10} 44.1 & \cellcolor{gray!10} 75.0 & \cellcolor{gray!10} 56.5 \\
\bottomrule[1.5pt]

\end{tabular}}
\caption{
Accuracy comparison of models across different control domains and interaction complexities (S-S, S-M, M-S, M-M) under Function Call (FC) and State-based Function Call (SFC) paradigms \textbf{with reflection}. Overall, SFC consistently outperforms FC across all domains and task types.
}
\vspace{-1.5em}
\label{tables:reflect-results}
\end{table*}
\begin{table*}[t]
\belowrulesep=0pt
\aboverulesep=0pt
\fontsize{14}{21}\selectfont
\centering

\resizebox{\textwidth}{!}{
\begin{tabular}{l c|cccc| cccc| cccc| cccc}
\toprule[1.5pt]
\multirow{2}{*}{\textbf{Model Name}}& \multirow{2}{*}{\textbf{Overall}}  & \multicolumn{4}{c|}{\textbf{Multimedia}}  & \multicolumn{4}{c|}{\textbf{Touch Control}}& \multicolumn{4}{c|}{\textbf{Car Control}}& \multicolumn{4}{c}{\textbf{Light}}\\
\cmidrule(lr){3-6} \cmidrule(lr){7-10} \cmidrule(lr){11-14} \cmidrule(lr){15-18}
&& \textbf{S-S} & \textbf{S-M} & \textbf{M-S} & \textbf{M-M} &\textbf{S-S} &\textbf{S-M} &\textbf{M-S} &\textbf{M-M} & \textbf{S-S} & \textbf{S-M} & \textbf{M-S} & \textbf{M-M} & \textbf{S-S} & \textbf{S-M} & \textbf{M-S} & \textbf{M-M}  \\
 \midrule

\hline \multicolumn{18}{c}{\textit{\textbf{Function Call}}} \\   \hline 

GPT-4o & \textbf{60.0} & \textbf{72.1} & \textbf{38.8} & \textbf{66.7} & \textbf{48.4} & \textbf{39.6} & \textbf{30.0} & \textbf{81.2} & \textbf{45.2} & \textbf{56.4} & \textbf{40.0} & \textbf{52.4} & \textbf{50.1} & \textbf{65.6} & \textbf{45.9} & \textbf{72.1} & \textbf{55.3} \\
DeepSeek-v3-250324 & 36.7 & 54.6 & 14.0 & 52.3 & 36.3 & 34.4 & 7.0 & 66.7 & 38.5 & 22.6 & 13.2 & 37.7 & 32.1 & 26.6 & 14.3 & 40.3 & 32.3 \\
Llama-3.1-8B & 25.4 & 50.0 & 11.7 & 36.9 & 25.7 & 21.9 & 2.9 & 49.6 & 15.5 & 14.3 & 9.2 & 24.2 & 18.0 & 7.8 & 8.1 & 26.5 & 19.5 \\
Qwen-2.5-72B & 27.6 & 52.9 & 11.7 & 41.0 & 26.6 & 25.0 & 2.9 & 53.4 & 18.5 & 15.8 & 9.2 & 26.6 & 20.0 & 9.1 & 8.1 & 29.9 & 21.4 \\
Qwen-2.5-32B & 27.6 & 51.8 & 13.0 & 39.0 & 31.7 & 21.9 & 2.9 & 52.8 & 20.2 & 17.3 & 9.2 & 27.4 & 19.9 & 15.6 & 8.1 & 28.3 & 21.0 \\
Qwen-2.5-14B & 29.3 & 51.1 & 11.7 & 43.5 & 28.4 & 21.9 & 2.9 & 53.1 & 16.3 & 15.0 & 9.2 & 27.2 & 22.0 & 16.9 & 9.7 & 34.8 & 23.8 \\
Qwen-2.5-7B & 28.0 & 57.1 & 12.0 & 39.9 & 28.9 & 29.7 & 2.9 & 53.3 & 19.9 & 15.0 & 9.3 & 25.1 & 23.3 & 11.7 & 8.3 & 28.7 & 25.6 \\
\cellcolor{gray!10} \textbf{Avg.} & \cellcolor{gray!10} 33.5 & \cellcolor{gray!10} 55.7 & \cellcolor{gray!10} 16.1 & \cellcolor{gray!10} 45.6 & \cellcolor{gray!10} 32.3 & \cellcolor{gray!10} 27.8 & \cellcolor{gray!10} 7.4 & \cellcolor{gray!10} 58.6 & \cellcolor{gray!10} 24.9 & \cellcolor{gray!10} 22.3 & \cellcolor{gray!10} 14.2 & \cellcolor{gray!10} 31.5 & \cellcolor{gray!10} 26.5 & \cellcolor{gray!10} 21.9 & \cellcolor{gray!10} 14.6 & \cellcolor{gray!10} 37.2 & \cellcolor{gray!10} 28.4 \\


\hline \multicolumn{18}{c}{\textit{\textbf{State-based Function Call}}} \\   \hline 

GPT-4o & 66.9 & \textbf{66.1} & 52.8 & \textbf{71.1} & 52.6 & 69.3 & 36.6 & 87.0 & 51.2 & 55.7 & 49.5 & 56.0 & 60.6 & 75.9 & 49.2 & 79.8 & 65.7 \\
DeepSeek-v3-250324 & \textbf{68.9} & 65.6 & \textbf{61.4} & 69.2 & \textbf{54.2} & \textbf{76.6} & \textbf{41.1} & \textbf{91.7} & \textbf{69.1} & \textbf{69.9} & \textbf{53.8} & \textbf{56.4} & \textbf{62.0} & 62.0 & \textbf{54.9} & \textbf{80.5} & \textbf{66.6} \\
Llama-3.1-8B & 42.3 & 55.9 & 31.2 & 50.8 & 31.6 & 42.2 & 9.2 & 69.5 & 29.6 & 30.3 & 19.3 & 29.8 & 25.4 & 44.3 & 20.9 & 57.3 & 29.8 \\
Qwen-2.5-72B & 57.9 & 63.7 & 42.4 & 61.2 & 47.6 & 62.4 & 33.2 & 79.8 & 43.5 & 37.7 & 37.0 & 44.6 & 49.9 & \textbf{77.2} & 45.7 & 77.0 & 55.4 \\
Qwen-2.5-32B & 51.6 & 61.9 & 35.7 & 57.8 & 46.1 & 39.6 & 10.1 & 69.3 & 32.9 & 28.6 & 21.1 & 38.4 & 48.5 & 70.3 & 35.1 & 74.3 & 55.1 \\
Qwen-2.5-14B & 28.6 & 50.8 & 15.1 & 37.9 & 28.3 & 23.4 & 2.9 & 50.2 & 18.1 & 22.4 & 10.7 & 24.6 & 19.0 & 10.1 & 8.3 & 36.3 & 21.7 \\
Qwen-2.5-7B & 35.9 & 52.5 & 18.9 & 41.1 & 29.0 & 28.1 & 8.0 & 56.8 & 21.7 & 26.1 & 12.0 & 26.1 & 22.2 & 44.3 & 19.0 & 52.7 & 27.1 \\
\cellcolor{gray!10} \textbf{Avg.} & \cellcolor{gray!10} 50.3 & \cellcolor{gray!10} 59.5 & \cellcolor{gray!10} 36.8 & \cellcolor{gray!10} 55.6 & \cellcolor{gray!10} 41.3 & \cellcolor{gray!10} 48.8 & \cellcolor{gray!10} 20.2 & \cellcolor{gray!10} 72.0 & \cellcolor{gray!10} 38.0 & \cellcolor{gray!10} 38.7 & \cellcolor{gray!10} 29.1 & \cellcolor{gray!10} 39.4 & \cellcolor{gray!10} 41.1 & \cellcolor{gray!10} 54.9 & \cellcolor{gray!10} 33.3 & \cellcolor{gray!10} 65.4 & \cellcolor{gray!10} 45.9 \\

\bottomrule[1.5pt]

\end{tabular}}
\caption{
Accuracy comparison of models across different control domains and interaction complexities (S-S, S-M, M-S, M-M) under Function Call (FC) and State-based Function Call (SFC) paradigms \textbf{without sample}. Overall, SFC consistently outperforms FC across all domains and task types.
}
\vspace{-1em}
\label{tables:nosample-results}
\end{table*}

We implement three prompting strategies that operate within two different paradigms: Function Call (FC) and State-based Function Call (SFC). Both paradigms use variants of the ReAct framework.

\paragraph{ReAct.} The core ReAct framework follows an iterative process where the agent first observes the current system state or results from previous API calls. Based on these observations, the agent reasons about what to do next and generates the corresponding executable code, either API calls or state transition code. In the FC paradigm, the agent generates API calls which are then executed by the VehicleWorld executor. The execution results are returned as feedback to guide the agent's next reasoning step. In the SFC paradigm, the agent produces code that directly modifies the JSON-formatted system state. This code is executed by the environment, which then returns both validation results and the updated system state as feedback.

\paragraph{ReAct + Reflection.} We also test ReAct with reflection by providing both paradigms with 3 additional reflection opportunities. During these reflection turns, agents can re-observe the system state, reason about whether their actions achieved the intended outcome, and take corrective actions if needed.

\paragraph{ReAct without Examples.} In both the standard ReAct and ReAct + reflection approaches, we provide demonstration examples to help agents understand task requirements and ensure basic instruction-following. To evaluate zero-shot performance, we implement a variant that removes these examples entirely, testing the model's ability to understand and execute tasks without explicit demonstrations.

\subsection{Error Rate} 
\label{app:error-rate}

To assess the reliability of different evaluation strategies, we manually annotated 200 Function Call (FC) outputs from the Qwen2.5 model series. Each output was labeled as correct or incorrect by multiple experts in the field of intelligent cockpit systems, based on whether it fulfilled the user intent expressed in the original query.

We then compared the results of two automatic evaluation methods, state-based and rule-based, with the expert annotations. The rule-based method checks the API call sequence for exact matches in API names, parameter keys and values, and call count. In contrast, the state-based method judges correctness based on whether the final system state satisfies the intended task goal.

The error rate is defined as the proportion of mismatches between the automatic evaluation result and the human annotation:
\[
\text{Error Rate} = \frac{\text{FP} + \text{FN}}{\text{TP} + \text{TN} + \text{FP} + \text{FN}}.
\]
This metric reflects how closely each automatic evaluation method aligns with human judgments of functional correctness. A lower error rate indicates that the evaluation strategy more accurately captures the user's intent and the practical outcome of the task.

\subsection{Evaluation Detail}
\label{app:evaluation-detail}

During evaluation, we prompt the model to sequentially respond to each query by either generating API calls (FC) or producing state transitions (SFC), forming a multi-turn interaction framework. The prompts used for evaluation are described in Appendix~\ref{app:prompts}. As illustrated in Figure~\ref{fig:benchmark} (right), after the agent generates a response, we extract the code enclosed within \verb|```python```| code blocks and execute it in a local environment.

For FC, the agent directly generates an API call, while in SFC, the agent outputs corresponding state transition code. After execution, we collect return values, exceptions, and logs as feedback and return them to the agent for the next turn. In SFC, we additionally append the current device states to the feedback after code execution.

The state-aware design in SFC introduces significantly longer input sequences due to the accumulation of system state in scenarios with multiple queries. To address potential input length issues, we implement a lightweight conversation management strategy: whenever a new query begins, all device states attached to previous turns are discarded, retaining only the latest states. This keeps the context window within reasonable bounds while preserving essential information required for accurate evaluation. 

\section{Metric Detail}
\label{app:metric-detail}
As described in Section~\ref{benchmark:metrics}, we employ three distinct metrics to evaluate FC and SFC, conducting state-based assessments of model performance.

\subsection{F1 positive}
This metric evaluates how effectively the model identifies attributes that require changes and performs modifications on them. It focuses on the act of modification rather than the correctness of the modified values.

\paragraph{Precision:} The proportion of correctly modified attributes among all attributes the model attempted to modify.
\begin{equation}
P_{positive} = \frac{TP}{TP + FP}
\end{equation}

\paragraph{Recall:} The proportion of attributes that should be modified and are actually modified by the model among all attributes that should be modified.
\begin{equation}
R_{positive} = \frac{TP}{total\_should\_changed}
\end{equation}

\paragraph{F1 Score:} The harmonic mean of precision and recall.
\begin{equation}
F1_{positive} = \frac{2 \cdot P_{positive} \cdot R_{positive}}{P_{positive} + R_{positive}}
\end{equation}

Here, $TP$ denotes the number of attributes that should be modified and are indeed modified by the model, $FP$ denotes the number of attributes that should not be modified but are modified by the model, and $total\_should\_changed$ represents the total number of attributes that should be modified.

\subsection{F1 negative}
This metric evaluates how effectively the model identifies attributes that should remain unchanged and preserves them without modification. It focuses on the act of preservation rather than the overall correctness of the system state.

\paragraph{Precision:} The proportion of correctly preserved attributes among all attributes the model predicted should remain unchanged.
\begin{equation}
P_{negative} = \frac{negative\_TP}{negative\_TP + negative\_FP}
\end{equation}

\paragraph{Recall:} The proportion of attributes that should remain unchanged and are actually preserved by the model among all attributes that should remain unchanged.
\begin{equation}
R_{negative} = \frac{negative\_TP}{total\_should\_unchanged}
\end{equation}

\paragraph{F1 Score:} The harmonic mean of precision and recall.
\begin{equation}
F1_{negative} = \frac{2 \cdot P_{negative} \cdot R_{negative}}{P_{negative} + R_{negative}}
\end{equation}

Here, $negative\_TP$ denotes the number of attributes that should be preserved and are indeed preserved by the model, $negative\_FP$ denotes the number of attributes that should be modified but are preserved by the model, and $total\_should\_unchanged$ represents the total number of attributes that should remain unchanged.

\subsection{Accuracy}
This metric quantifies the correctness of modification values, measuring how often the model assigns the correct new values to attributes that need to be changed. Unlike $F1_{positive}$ and $F1_{negative}$ which focus on modification behaviors, this metric evaluates the actual correctness of the modified values.

\begin{equation}
acc = \frac{N_{correct}}{N_{total}}
\end{equation}

Where $N_{correct}$ represents the number of attributes that were both modified and assigned the correct target values, and $N_{total}$ denotes the total number of attributes requiring modification. 

Moreover, for numerical attributes, exact value comparison is inappropriate when user requests are ambiguous (e.g., "increase the volume" without specifying by how much). In these cases, we evaluate accuracy based on the trend (increase, decrease, or maintain) rather than requiring exact value matches. This approach better aligns with user intent when precise numerical targets aren't explicitly stated in the request.

Together, these three metrics provide a comprehensive evaluation framework: $F1_{positive}$ measures the model's ability to identify and modify the right attributes, $F1_{negative}$ assesses its ability to preserve unchanged attributes, and Accuracy evaluates the correctness of the actual modification values.

\begin{table*}[t]
\centering
\footnotesize
\begin{tabular}{|l|c|c|c|c|c|c|}
\hline
\textbf{Models} & \textbf{\# Para} & \textbf{Launch Time} & \textbf{Max Tokens} & \textbf{Scaling} & \textbf{Corporation} & \textbf{License} \\
\hline
GPT-4o & / & May 13, 2024 & 128,000 & Effort & OpenAI & Proprietary \\
\hline
Claude-3.7-Sonnet & / & Feb 24, 2025 & 200,000 & Budget & Anthropic & Proprietary \\
\hline
DeepSeek-v3-250324 & 671B & Mar 25, 2025 & 131,072 & Budget & DeepSeek & Open Source \\
\hline
Llama-3.1-8B & 8B & Jul 23, 2024 & 131,072 & Budget & Meta & Llama License \\
\hline
Qwen-2.5-7B & 7.6B & Sep 19, 2024 & 131,072 & Budget & Alibaba & Apache 2.0 \\
\hline
Qwen-2.5-14B & 14.7B & Sep 19, 2024 & 131,072 & Budget & Alibaba & Apache 2.0 \\
\hline
Qwen-2.5-32B & 32.5B & Sep 19, 2024 & 131,072 & Budget & Alibaba & Apache 2.0 \\
\hline
Qwen-2.5-72B & 72.7B & Sep 19, 2024 & 131,072 & Budget & Alibaba & Apache 2.0 \\
\hline
Qwen-3-4B & 4.0B & Apr 29, 2025 & 131,072 & Budget & Alibaba & Apache 2.0 \\
\hline
Qwen-3-8B & 8.2B & Apr 29, 2025 & 131,072 & Budget & Alibaba & Apache 2.0 \\
\hline
Qwen-3-32B & 32.8B & Apr 29, 2025 & 131,072 & Budget & Alibaba & Apache 2.0 \\
\hline
\end{tabular}
\caption{Large language models evaluated in our experiments with specifications and characteristics.}
\label{tables:model-detail}
\vspace{-1em}
\end{table*}

\section{Dataset Detail}
\label{app:dataset-license}
The dataset used in VehicleWorld is constructed entirely by the authors without incorporating any third-party content, personal data, or system logs. To prevent any privacy leakage, we employed synthetic data and rigorously ensured that the generated data contained no personal information or offensive content. We measured the distribution against participants' real intentions using Jensen-Shannon divergence and obtained a similarity score of 0.9640, demonstrating strong consistency. The Jensen-Shannon divergence is defined as:
\begin{equation}
JSD(P | Q) = \frac{1}{2} D_{KL}(P | M) + \frac{1}{2} D_{KL}(Q | M)
\end{equation}
where \( P \) and \( Q \) represent the empirical distributions of the synthetic data and real-world participant intentions, respectively, and \( M = \frac{1}{2}(P + Q) \) is the average distribution. \( D_{\mathrm{KL}} \) denotes the Kullback–Leibler divergence. 

Throughout our construction process, we informed all participants of our intended usage and ensured that our methods strictly adhered to our mutual agreement. As detailed in Appendix~\ref{sec:instruction}, we present the specific instructions provided to annotators in Table~\ref{tab:api-design-instruction}, \ref{tab:test-set-instruction}, and \ref{tab:function-call-evaluation}. The objectives of our data collection and utilization are outlined in Table~\ref{tab:annotation-purpose}. Our dataset is in English, with participants from the Pacific Rim region. During collection, we considered diversity in gender and ethnicity. The dataset will be released under the Apache License 2.0, and we strongly advocate that all related usage strictly complies with relevant regulations.

\section{Model Detail}
\label{app:model-detail}
Our evaluation includes both closed-source and open-source language models. For closed-source models, we selected GPT-4o~\cite{OpenAI2024GPT4o} and Anthropic's Claude-3.7-Sonnet~\cite{Anthropic2025Claude37Sonnet}. For open-source models, we evaluated DeepSeek-v3~\cite{liu2024deepseek}, Llama-3.1-8B-Instruct~\cite{grattafiori2024llama}, and a range of models from the Qwen family. This includes the classic Qwen-2.5 series (7B, 14B, 32B, 72B)~\cite{qwen2.5} , as well as Qwen's latest models: Qwen-3 series (4B, 8B, 32B)~\cite{qwen3}. For detailed information about the models used in our experiments, please refer to Table~\ref{tables:model-detail}.


\section{Class Detail}
\subsection{Prompt}
\label{app:class-prompt}
The comprehensive prompts used for class construction are meticulously detailed in Appendix \ref{app:prompts}.
\subsection{Class Architecture}
\label{app:class}
The VehicleWorld framework uses a modular class architecture where each vehicle device is implemented as an independent module class. Each device class contains specialized inner classes for subcomponents, creating a clean hierarchical structure.
To represent device states in a readable format, each class implements a \texttt{to\_dict()} method that organizes instance attributes into structured JSON documents with detailed metadata. These methods work recursively: the \texttt{to\_dict()} method of each top-level device class automatically calls the \texttt{to\_dict()} methods of its nested submodules to generate a complete, hierarchical state representation.

We also implement a global static \texttt{Environment} class that manages system-wide properties such as temperature settings, audio volume levels, and communication channels, ensuring consistent behavior across all devices and reduces code redundancy.

Finally, each device class provides multiple initialization methods through specialized \texttt{init()} class methods. These allow flexible configuration of different initial states for testing and demonstration purposes by combining different initialization approaches for specific scenarios. The following example demonstrates this architecture through the Navigation device class implementation:

\DeclareFixedFont{\ttb}{T1}{txtt}{bx}{n}{9} 
\DeclareFixedFont{\ttm}{T1}{txtt}{m}{n}{9}  
\definecolor{codedeepblue}{rgb}{0,0,0.5}
\definecolor{codedeepred}{rgb}{0.6,0,0}
\definecolor{codedeepgreen}{rgb}{0,0.5,0}
\definecolor{codegray}{rgb}{0.9,0.9,0.9}
\definecolor{backcolour}{rgb}{0.95,0.95,0.95}
\providecommand{\listingsttfamily}{}  
\renewcommand{\listingsttfamily}{\fontfamily{IBMPlexMono-TLF}\small}

\lstdefinestyle{mypython}{ 
  backgroundcolor=\color{backcolour},  
  basicstyle=\footnotesize\ttfamily,       
  breakatwhitespace=false, 
  breaklines=true,              
  captionpos=b,                 
  commentstyle=\color{codedeepred}\textit,    
  frame=tb,	                   	
  keepspaces=true,                
  keywordstyle=\color{codedeepblue}\bfseries, 
  language=Python,
  otherkeywords={*,...},    
  numbers=none,  
  numbersep=5pt,              
  numberstyle=\tiny\color{commentsColor}, 
  rulecolor=\color{white},        
  showspaces=false,                
  showstringspaces=false,          
  showtabs=false,                  
  stepnumber=1,                   
  stringstyle=\color{codedeepgreen},
  tabsize=2
}

\providecommand{\yamlvalue}[1]{{\textcolor{black}{#1}}}
\lstdefinestyle{myyaml}{
     basicstyle=\color{blue}\footnotesize\listingsttfamily,
     rulecolor=\color{black},
     stringstyle=\color{blue},
     comment=[l]{:},
     commentstyle=\color{black},
     showspaces=false,                
    showstringspaces=false,          
    showtabs=false,
    breaklines=true,
    escapeinside={<@}{@>}
 }

\lstdefinestyle{mytext}{ 
  backgroundcolor=\color{backcolour},  
  basicstyle=\footnotesize\ttfamily,    
     morekeywords={SYSTEM, USER, ASSISTANT, TOOL, SEQUENCE, HIDDEN},
     keywordstyle=\color{black}\bfseries\underbar,
     comment=[s]{```python}{```},
     commentstyle=\color{blue},
  breakatwhitespace=false, 
  breaklines=true,              
  captionpos=b,                 
  frame=tb,	                   	
  keepspaces=true,                
  rulecolor=\color{white},        
  showspaces=false,                
  showstringspaces=false,          
  showtabs=false,                  
  stepnumber=1,                   
  tabsize=2
}
\onecolumn
\begin{lstlisting}[style=mypython,label={list:navigation-class},caption={Entity class for navigation device. This class demonstrates the core design of device abstraction, including inner classes, property encapsulation, and API method implementation. Some content has been omitted for brevity},captionpos=t]
import Environment
class Navigation:
    class DistanceUnit(Enum):
        """
        Enumeration for distance units used in navigation.
        """
        KILOMETERS = "Kilometers"
        MILES = "Miles"
    
    class VolumeLevel(Enum):
        """
        Enumeration for predefined volume levels.
        """
        MAX = "max"
        HIGH = "high"
        MEDIUM = "medium"
        LOW = "low"
        MIN = "min"
        
    class RouteInfo:
        """
        Inner class for storing route information between two points.
        """
        def __init__(self, departure="Current location", destination=""):
            self._departure = departure
            self._destination = destination
        
        @property
        def departure(self):
            return self._departure
            
        @departure.setter
        def departure(self, value):
            self._departure = value
            
        @property
        def destination(self):
            return self._destination
            
        @destination.setter
        def destination(self, value):
            self._destination = value
            
        def to_dict(self):
            """
            Convert RouteInfo object to a dictionary with metadata.
            
            Returns:
                dict: Dictionary representation of route information.
            """
            return {
                "departure": {
                    "value": self.departure,
                    "description": "Starting point of the route",
                    "type": type(self.departure).__name__
                },
                "destination": {
                    "value": self.destination,
                    "description": "End point of the route",
                    "type": type(self.destination).__name__
                }
            }
    
    def __init__(self):
        # Base navigation state
        self._is_active = False
        # Route information
        self._current_route = self.RouteInfo()
    
    # Property getters and setters
    @property
    def is_active(self):
        """Get navigation active status."""
        return self._is_active
    
    @is_active.setter
    def is_active(self, value):
        """Set navigation active status."""
        self._is_active = value
    
    @property
    def current_route(self):
        """Get current navigation route."""
        return self._current_route
    
    @current_route.setter
    def current_route(self, value):
        """Set current navigation route."""
        self._current_route = value

    # API Methods Implementation
    @api("navigation")
    def navigation_meter_unit(self, mode):
        """
        Set the unit of distance displayed on the navigation.
        
        Args:
            mode (str): Distance unit options. Valid values are "Kilometers" or "Miles".
        
        Returns:
            dict: Result of the operation with updated unit.
        """
        try:
            # Update the global environment unit system
            if mode == "Kilometers":
                Environment.set_unit_system("kilometer")
            else:
                Environment.set_unit_system("mile")
            return {
                "success": True,
                "unit": mode
            }
        except ValueError as e:
            return {
                "success": False,
                "error": str(e)
            }
    
    @api("navigation")
    def navigation_soundVolume_set(self, value=None, degree=None):
        """
        Set navigation volume adjustment.
        
        Args:
            value (int, optional): Numeric volume value (0-100).
            degree (str, optional): Categorical volume level. 
                                    Valid values are "max", "high", "medium", "low", "min",
                                    mutually exclusive with value.
        
        Returns:
            dict: Result of the operation with updated volume.
        """
        if value is not None and degree is not None:
            return {
                "success": False,
                "error": "Cannot specify both value and degree"
            }
        
        try:
            if value is not None:
                if not isinstance(value, int) or not (0 <= value <= 100):
                    return {
                        "success": False,
                        "error": "Volume value must be an integer between 0 and 100"
                    }
                Environment.set_volume(value)
            elif degree is not None:
                # Map degree to volume value
                degree_map = {
                    self.VolumeLevel.MAX.value: 100,
                    self.VolumeLevel.HIGH.value: 80,
                    self.VolumeLevel.MEDIUM.value: 50,
                    self.VolumeLevel.LOW.value: 20,
                    self.VolumeLevel.MIN.value: 0
                }
                
                if degree not in degree_map:
                    return {
                        "success": False,
                        "error": f"Invalid volume degree: {degree}. Valid values are {list(degree_map.keys())}"
                    }
                
                Environment.set_volume(degree_map[degree])
            else:
                return {
                    "success": False,
                    "error": "Either value or degree must be specified"
                }
            
            # Set global environment volume and change sound channel to navigation
            Environment.set_sound_channel("navigation")
            
            return {
                "success": True,
                "volume": Environment.get_volume(),
                "sound_channel": Environment.get_sound_channel()
            }
        except Exception as e:
            return {
                "success": False,
                "error": str(e)
            }
    
    @api("navigation")
    def navigation_get_destination(self):
        """
        Get current navigation destination.
        
        Returns:
            dict: Information about the current destination.
        """
        if not self.is_active or self.current_route is None:
            return {
                "success": False,
                "error": "No active navigation route"
            }
        
        return {
            "success": True,
            "destinationInfo": self.current_route.destination
        }
    
    @api("navigation")
    def navigation_exit(self):
        """
        Exit navigation by deactivating the current route.
        
        Returns:
            dict: Result of the operation.
        """
        if not self.is_active:
            return {
                "success": False,
                "error": "Navigation is not active"
            }
        
        self.is_active = False
        self.current_route = None
        self.waypoints = []
        
        return {
            "success": True,
            "message": "Navigation exited successfully"
        }
    
    @api("navigation")
    def navigation_route_plan(self, address, placeOfDeparture="Current location"):
        """
        Route planning, specify destination to start navigation.
        
        Args:
            address (str): Destination name/address.
            placeOfDeparture (str): Departure name/address. Defaults to "Current location".
            
        Returns:
            dict: Result of the operation with route details.
        """
        if not address:
            return {
                "success": False,
                "error": "Destination address is required"
            }
        
        # Create a new route with the specified destination
        self.current_route = self.RouteInfo(destination=address, departure=placeOfDeparture)
        self.is_active = True
        self.waypoints = []
        
        # Set sound channel to navigation for voice guidance
        Environment.set_sound_channel("navigation")
        
        return {
            "success": True,
            "route": self.current_route.to_dict()
        }

    def to_dict(self):
        """
        Convert the Navigation object to a dictionary with metadata.
        
        Returns:
            dict: Dictionary representation of the Navigation object.
        """
        waypoints_info = []
        for i, waypoint in enumerate(self.waypoints):
            waypoints_info.append({
                "index": i,
                "location": waypoint
            })
        
        return {
            "is_active": {
                "value": self.is_active,
                "description": "Whether navigation is currently active, when you need to use navigation, you should set it to True.",
                "type": type(self.is_active).__name__
            },
            "current_route": {
                "value": self.current_route.to_dict() if self.current_route else None,
                "description": "Current active navigation route. Set it to None when exit navigation",
                "type": "RouteInfo or None"
            }
        }

    @classmethod
    def init1(cls):
        """
        Initialize a Navigation instance with active navigation to Shanghai.
        
        Returns:
            Navigation: An initialized Navigation instance with active navigation to Shanghai.
        """
        instance = cls()
        # Set navigation to active
        Environment.set_sound_channel("navigation")
        instance.is_active = True
        # Set up the current route
        instance.current_route = cls.RouteInfo(departure="Current location", destination="Shanghai")  
        return instance

    @classmethod
    def init2(cls):
        """
        Initialize a Navigation instance with navigation not active.
        
        Returns:
            Navigation: An initialized Navigation instance with inactive navigation.
        """
        instance = cls()
        # Ensure navigation is inactive
        instance.is_active = False
        # Reset current route
        instance.current_route = cls.RouteInfo()
        return instance
\end{lstlisting}

\subsection{Global Static Environment Class}
\label{app:environment}

\onecolumn
\begin{lstlisting}[style=mypython,label={list:navigation-class},caption={Implementation of the global \texttt{Environment} class as a centralized resource manager. The class follows a simplified Singleton design pattern and manages shared cockpit attributes such as volume, sound channel, unit system, temperature, and time format. All device entity classes interact with this environment interface to ensure consistency and avoid redundant state maintenance.},captionpos=t]
class Environment:
    _context = {
        "volume": 50,
        "sound_channel": "music",
        "unit_system": "mile",
        "timestamp": "2025-04-13 12:00:00",
        "speaker": "driver's seat",
        "temperature": 14,
        "language": "Chinese",
        "time_display_format": "24-hour-format"
    }

    @classmethod
    def set_language(cls, language): cls._context["language"] = language
    @classmethod
    def get_language(cls): return cls._context["language"]

    @classmethod
    def set_time_display_format(cls, time_format): cls._context["time_display_format"] = time_format
    @classmethod
    def get_time_display_format(cls): return cls._context["time_display_format"]

    @classmethod
    def get_current_speaker(cls): return cls._context["speaker"]

    @classmethod
    def set_temperature(cls, temperature): cls._context["temperature"] = temperature
    @classmethod
    def get_temperature(cls): return cls._context["temperature"]

    @classmethod
    def get_timestamp(cls): return cls._context["timestamp"]
    @classmethod
    def set_timestamp(cls, timestamp): cls._context["timestamp"] = timestamp

    @classmethod
    def set_volume(cls, volume): cls._context["volume"] = volume
    @classmethod
    def get_volume(cls): return cls._context["volume"]

    @classmethod
    def set_sound_channel(cls, channel): cls._context["sound_channel"] = channel
    @classmethod
    def get_sound_channel(cls): return cls._context["sound_channel"]

    @classmethod
    def set_unit_system(cls, unit): cls._context["unit_system"] = unit
    @classmethod
    def get_unit_system(cls): return cls._context["unit_system"]

    @classmethod
    def to_dict(cls) -> Dict[str, Any]:
        return {
            "volume": {
                "type": "int",
                "value": cls._context["volume"],
                "description": "Volume level (0-100)"
            },
            "sound_channel": {
                "type": "str",
                "value": cls._context["sound_channel"],
                "description": "..."
            },
            "unit_system": {
                "type": "str",
                "value": cls._context["unit_system"],
                "description": "Distance unit system, supports mile or kilometer"
            },
            "timestamp": {
                "type": "str",
                "value": cls._context["timestamp"],
                "description": "Current system time"
            },
            "speaker": {
                "type": "str",
                "value": cls._context["speaker"],
                "description": "..."
            },
            "temperature": {
                "type": "int",
                "value": cls._context["temperature"],
                "description": "..."
            },
            "language": {
                "type": "str",
                "value": cls._context["language"],
                "description": "Current system language"
            },
            "time_display_format": {
                "type": "str",
                "value": cls._context["time_display_format"],
                "description": "Time format, either '24-hour-format' or '12-hour-format'"
            }
        }

    @classmethod
    def init1(cls):
        cls._context.update({
            "volume": 60,
            "sound_channel": "music",
            "unit_system": "mile",
            "timestamp": "2025-04-13 11:00:00",
            "speaker": "driver's seat",
            "temperature": 14,
            "language": "Chinese",
            "time_display_format": "24-hour-format"
        })
        return cls

    @classmethod
    def init2(cls):
        cls._context.update({
            "volume": 75,
            "sound_channel": "video",
            "unit_system": "mile",
            "timestamp": "2025-04-13 12:10:00",
            "speaker": "passenger seat",
            "temperature": 16,
            "language": "Chinese",
            "time_display_format": "24-hour-format"
        })
        return cls
\end{lstlisting}

\section{Prompts}
\label{app:prompts}
\begin{lstlisting}[
  style=mytext,
  label={list:class-construct},
  caption={Structured prompt template used with Claude-3.7-Sonnet to systematically generate VehicleWorld entity classes through a four-stage process: API analysis, entity class implementation with getters/setters, complete API method implementation with error handling, and thorough documentation via to\_dict() methods.},
  captionpos=t
]
# {Device} System API Analysis and Entity Class Implementation

## Context
You are in a vehicle driving environment, and you are given a series of APIs related to the {Device} system, where {Device} is responsible for managing the in-car {Device} system.

## Task
Please analyze these APIs in detail, extract all implicit data attributes, and write a complete {Device} entity class using Python code.

## Requirements

### API Analysis
- Carefully analyze each API to identify the core data objects and states it operates on
- Extract all necessary attributes (fields) to support the functionality of these APIs
- Consider the relationships and dependencies between attributes
- Set reasonable initial values for each attribute
- Ensure that each API call is reflected in changes to one or more attributes

### Entity Class Implementation
- Implement getter and setter methods for all attributes
- If the top-level {Device} entity class requires complex data structures as attributes, you can define necessary inner classes

### API Method Implementation
Design separate method implementations for each API, with each method requiring:
1. Functionality implemented through executing relevant getter and setter methods
2. Clear method signatures, including parameter types and return types
3. Detailed parameter validation and error handling
4. Clear state transition logic
5. Structured return values, including operation results and related state information
6. Add @api("{Device}") annotation for each API method
7. If API method parameters involve enumeration classes, the possible values should be specified in the comments

### Documentation
- Each entity class needs to set up a to_dict method to print the class's attributes, value types, and attribute descriptions
- If it's an enum attribute, the complete range of possible values needs to be given in the description

### Global Environment
- You can use a static global Environment class to get the current state of the environment, which is shared by multiple entity classes
- Some APIs need to call methods of the global class to modify the environment state

For example, {examples}

Here is an example of a class:
```python
from utils import api

class MyClass:
    class InnerExample:
        """
        Inner class, used as an attribute of MyClass, the example contains a simple attribute info.
        """
        def __init__(self, info):
            self.info = info

        def to_dict(self):
            return {
                "info": self.info,
                "description": "Example inner class information",
                "type": type(self.info).__name__
            }
        
        @classmethod
        def from_dict(cls, data):
            return cls(data["info"])

    def __init__(self, a, b, c, inner_info):
        self._a = a
        self._b = b
        self._c = c
        # Use an instance of the inner class as an attribute of MyClass
        self.inner = MyClass.InnerExample(inner_info)

    # Getter and setter for attribute a
    @property
    def a(self):
        return self._a

    @a.setter
    def a(self, value):
        self._a = value

    # Getter and setter for attribute b
    @property
    def b(self):
        return self._b

    @b.setter
    def b(self, value):
        self._b = value

    # Getter and setter for attribute c
    @property
    def c(self):
        return self._c

    @c.setter
    def c(self, value):
        self._c = value

    def to_dict(self):
        return {
            "a": {
                "value": self.a,
                "description": "Description of attribute a",
                "type": type(self.a).__name__
            },
            "b": {
                "value": self.b,
                "description": "Description of attribute b",
                "type": type(self.b).__name__
            },
            "c": {
                "value": self.c,
                "description": "Description of attribute c",
                "type": type(self.c).__name__
            },
            "inner": {
                "value": self.inner.to_dict(),
                "description": "Instance of inner class InnerExample",
                "type": "InnerExample"
            }
        }
    
    @classmethod
    def from_dict(cls, data):
        """
        Restore an instance of MyClass from dictionary data, including restoring inner class attributes.
        """
        a = data["a"]["value"]
        b = data["b"]["value"]
        c = data["c"]["value"]
        # First restore the inner class instance
        inner = MyClass.InnerExample.from_dict(data["inner"]["value"])
        instance = cls(a, b, c, inner.info)
        # If further synchronization of other attributes of the inner class is needed, it can be extended here
        instance.inner = inner
        return instance
    
    @api("MyClass")
    def api_function(self, a, b, c, inner_info):
        # Example implementation: update own attributes and inner class attributes
        self.a = a
        self.b = b
        self.c = c
        self.inner.info = inner_info
        return {
            "success": True,
            "updated_values": self.to_dict()
        }
```

Please provide a complete code implementation, below are the APIs provided to you:
{API document}
\end{lstlisting}

\begin{lstlisting}[
  style=mytext,
  label={list:class-construct},
  caption={Structured prompt template used with Claude-3.7-Sonnet to systematically generate VehicleWorld benchmark scenarios through HTML-style formatting: device initialization (\texttt{<inits>}), user queries (\texttt{<query>}), and corresponding API calls (\texttt{<api\_call>}) with automatic execution validation and expert verification for semantic correctness.},
  captionpos=t
]
# System Role
You are an in-vehicle artificial intelligence system with the following function modules:
{module_des}

# Main Task
Based on the current status of the function modules, design 10 different user query scenarios and generate corresponding API call chains for each scenario.

# Detailed Requirements
## Scenario Design
- If there are multiple modules, each scenario must involve at least 2 different modules
- Queries must have logical dependencies between them
- Scenarios must be based on the initial state and cannot violate it
- Scenarios must conform to real driving scenarios and user habits
- Use natural, conversational language for user queries

## API Call Constraints
- Clearly show the dependencies and calling relationships between APIs
- API calls need to reflect contextual coherence

## Output Format
Each scenario should use the following strict format:
```xml
<scenario>
    <query>User query 1 (natural language)</query>
    <api_call>vw.module.api0(parameter1, parameter2...)
vw.module.api1(parameter1, parameter2...)</api_call>(Use line break to separate multiple API calls)
    <query>User query 2 (natural language)</query>
    <api_call>vw.module.api2(parameter1, parameter2...)</api_call>
    <query>User query 3 (if any)</query>
    <api_call>vw.module.api3(parameter1, parameter2...) (if any)</api_call>
</scenario>
```
## Currrent World State
The current status of the function modules is as follows:
{State}

## API
Below you will be provided with information about the APIs for these modules:
{API documents}
\end{lstlisting}

\begin{lstlisting}[
  style=mytext, 
  label={list:evaluation},
  caption={FC evaluation prompt that guides agent through a structured API interaction process: module discovery, API querying, function execution, and feedback handling, with step-by-step instructions and example dialogues to ensure efficient task completion via API calls.},
  captionpos=t
]
You are an intelligent in-car AI assistant responsible for fulfilling user requests by calling the vehicle system API. Please ensure you understand user needs and complete tasks in the most efficient way.

**Step-by-Step Guidance**  
Please follow these steps sequentially:
1. **Get System Function Modules**  
   First call `Environment.search_module()` to obtain all the functionality modules of the vehicle system.

2. **Query APIs for Specific Modules**  
   After selecting the required function module, call `Environment.search_api(modules=['module_name'])` to query the API methods of that module, and then proceed with the calls.

3. **Get System Global Status (if needed)**  
   If the task involves global system environment (such as temperature, volume, etc.), call `Environment.to_dict()` to get the current system status.

**API Call Requirements**  
- Each code block can contain multiple related API calls, but should remain concise.
- Ensure the code can be executed directly, without non-code content other than comments.
- Code sections should be surrounded by ```python```.
- Only generate API calls, do not generate any additional code.

**System Feedback Handling**  
- The system will return execution results after each API call, including return values (success or failure) and any error messages.
- Carefully analyze each feedback to decide the next step. If an API call fails, analyze the reason and make corrections.

**Task Completion Standards**  
1. Choose the most efficient API combination to complete tasks, avoiding unnecessary code.
2. Correctly handle API errors and exception situations.
3. If the system status does not support the user's request, clearly decline.
**Example Task Flow**
User:
Navigate to the nearest gas station and turn up the navigation volume a bit

Assistant:
I'll help you navigate to the nearest gas station and increase the navigation volume. First I need to check the available system modules:
```python
Environment.search_module()
```

User:
Code execution results:
{{
  "modules": ["navigation", "music", "climate", "radio"]
}}

Assistant:
Now I'll query the API methods of the navigation module:
```python
Environment.search_api(modules=["navigation"])
```

User:
Code execution results:
{{
  "navigation": [
    "navigation_route_plan",
    "navigation_exit",
    "navigation_soundVolume_increase",
    "navigation_soundVolume_decrease",
    "navigation_soundVolume_set"
  ]
}}

Assistant:
After understanding the available APIs, I'll plan a route to the nearest gas station for you:
```python
vw.navigation.navigation_route_plan("nearest gas station")
```

User:
Code execution results:
{{
  "success": True,
  "route": {{
    "destination": "nearest gas station",
    "distance": "3 km", 
    "duration": "10 minutes"
  }}
}}

Assistant:
Navigation has been successfully set to the nearest gas station, now I'll increase the navigation volume:
```python
vw.navigation.navigation_soundVolume_increase(degree="little")
```

User:
Code execution results:
{{
  "success": True,
  "old_volume": 30,
  "new_volume": 40,
  "sound_channel": "navigation"
}}

Assistant:
I have successfully completed the following operations for you:
Set navigation to the nearest gas station, approximately 3 kilometers away, with an estimated travel time of 10 minutes
Increased the navigation volume from 30 to 40, making it easier for you to hear the navigation instructions clearly
The navigation system is now active, and you can follow the guidance to the gas station. If you need any other assistance, please let me know.
\end{lstlisting}

\begin{lstlisting}[
  style=mytext, 
  label={list:SFC-get-module},
  caption={SFC prompt for retrieving relevant modules, which are then passed to Listing~\ref{list:SFC-evaluation} to generate state transition code.},
  captionpos=t
]
You are a vehicle AI assistant. Given the current state of in-vehicle devices and a user query related to these devices, you need to identify and select the relevant device modules that should handle the query.

## Task Requirements:
- Analyze the user query and current system status
- Select the most relevant device modules (maximum 6 modules per query)
- Return the selected modules in the specified format

## Output Format:
<modules>
module1
module2
module3
</modules>

## Guidelines:
1. Each module name should match exactly with the keys in the system status dictionary
2. Separate multiple modules with newlines
3. Only select modules that are directly relevant to fulfilling the user's request
4. Consider the current state and capabilities of each module
5. Prioritize modules that are most likely to be needed for the task

## Example:

User:
Play the video I've downloaded
Current System Status:
{
    "video": {
        "value": {
            "current_video": {
                "value": {
                    "video_id": {
                        "value": "dl_001",
                        "type": "str",
                        "description": "Unique identifier for video"
                    },
                    "title": {
                        "value": "Highway Safety Tutorial",
                        "type": "str",
                        "description": "Title of the video"
                    },
                    "description": {
                        "value": "Learn about best practices for highway driving and safety tips",
                        "type": "str",
                        "description": "Description of the video"
                    }
                },
                "type": "VideoItem",
                "description": "Currently selected video"
            },
            "downloaded_videos": {
                "value": [
                    {
                        "video_id": {
                            "value": "dl_001",
                            "type": "str",
                            "description": "Unique identifier for video"
                        },
                        "title": {
                            "value": "City Night Drive",
                            "type": "str",
                            "description": "Title of the video"
                        },
                        "description": {
                            "value": "Exploring the city streets at night with ambient lighting",
                            "type": "str",
                            "description": "Description of the video"
                        }
                    }
                ],
                "type": "List[VideoItem]",
                "description": "List of downloaded videos"
            }
        },
        "description": "Video system for media playback",
        "type": "Video"
    },
    "radio": {
        "value": {
            "_history": {
                "type": "List[RadioStation]",
                "value": [
                    {
                        "name": {
                            "type": "str",
                            "value": "Indie Music Channel",
                            "description": "Name of the radio station"
                        },
                        "frequency_value": {
                            "type": "str",
                            "value": "90.5 MHz",
                            "description": "Frequency value of the radio station"
                        },
                        "city": {
                            "type": "str",
                            "value": "Portland",
                            "description": "City where the radio station is available"
                        },
                        "app_name": {
                            "type": "str",
                            "value": "Independent Music",
                            "description": "App name used to play this radio station"
                        },
                        "timestamp": {
                            "type": "float",
                            "value": "2025-04-13 11:00:00",
                            "description": "Timestamp when this station was last played"
                        }
                    }
                ],
                "description": "History of played radio stations (most recent first)"
            },
            "_collection": {
                "type": "List[RadioStation]",
                "value": [
                    {
                        "name": {
                            "type": "str",
                            "value": "Indie Music Channel",
                            "description": "Name of the radio station"
                        },
                        "frequency_value": {
                            "type": "str",
                            "value": "90.5 MHz",
                            "description": "Frequency value of the radio station"
                        },
                        "city": {
                            "type": "str",
                            "value": "Portland",
                            "description": "City where the radio station is available"
                        },
                        "app_name": {
                            "type": "str",
                            "value": "Independent Music",
                            "description": "App name used to play this radio station"
                        },
                        "timestamp": {
                            "type": "float",
                            "value": "2025-04-13 11:00:00",
                            "description": "Timestamp when this station was last played"
                        }
                    }
                ],
                "description": "Collection of favorite radio stations"
            }
        },
        "description": "Radio system for audio streaming",
        "type": "Radio"
    },
    "navigation": {
        "value": {
            "current_location": {
                "value": {
                    "latitude": 45.5152,
                    "longitude": -122.6784,
                    "address": "Portland, OR"
                },
                "type": "Location",
                "description": "Current vehicle location"
            },
            "destination": {
                "value": null,
                "type": "Location",
                "description": "Current navigation destination"
            }
        },
        "description": "Navigation and GPS system",
        "type": "Navigation"
    }
    ...
}

Assistant:
The user wants to play a downloaded video. The video module contains downloaded videos and current video information, making it the primary module needed to handle this request. The radio and navigation modules are not relevant for this specific task.
<modules>
video
</modules>
\end{lstlisting}

\begin{lstlisting}[
  style=mytext, 
  label={list:SFC-evaluation},
  caption={SFC evaluation prompt that guides agent through direct system state manipulation, teaching it to analyze the current vehicle state, generate appropriate state transition code, and verify results through updated status feedback, with clear response format requirements and illustrative examples.},
  captionpos=t
]
You are an intelligent vehicle AI assistant, your task is to help users analyze the vehicle system status to complete various tasks.

## Current System Status
1. The system will first provide you with the current status of the vehicle environment, including various information related to the task. You should fully understand the system status information and generate status modification code to complete the user-specified task.
2. After each code execution, the system status will be updated, and you need to determine whether the task has been completed or further operations are needed based on the return value of the code execution and the updated system status.
3. If the system status does not support executing the user's request, please refuse.

## Status Analysis Principles
When analyzing system status, please follow these principles:
1. Carefully check all available system modules and parameters
2. Understand the data types and value ranges of each field
3. Pay attention to the dependency relationships of status values, such as certain operations requiring specific sound channels
4. Prioritize using the ready-made data provided by the system, avoiding creating unnecessary new values
5. Ensure changes comply with system constraints

## Response Format
Your response should include the following parts:
1. Brief analysis of the user's question and code execution results
2. Code execution section (surrounded by ```python```, containing only status changes, do not include other code)
3. Do not generate other additional content

## Example Task Flow

### Example 1
User:
Play the video I've downloaded
Current system status:
vw = {
    "video": {
                "value": {
                    "current_video": {
                        "value": {
                            "video_id": {
                                "value": "dl_001",
                                "type": "str",
                                "description": "Unique identifier for video"
                            },
                            "title": {
                                "value": "Highway Safety Tutorial",
                                "type": "str",
                                "description": "Title of the video"
                            },
                            "description": {
                                "value": "Learn about best practices for highway driving and safety tips",
                                "type": "str",
                                "description": "Description of the video"
                            }
                            ...
                        },
                        "type": "VideoItem",
                        "description": "Currently selected video."
                    },
                    "downloaded_videos": {
                        "value": [
                            {
                                "video_id": {
                                    "value": "dl_001",
                                    "type": "str",
                                    "description": "Unique identifier for video"
                                },
                                "title": {
                                    "value": "City Night Drive",
                                    "type": "str",
                                    "description": "Title of the video"
                                },
                                "description": {
                                    "value": "Exploring the city streets at night with ambient lighting",
                                    "type": "str",
                                    "description": "Description of the video"
                                }
                                ...
                            },
                        ],
                        "type": "List[VideoItem]",
                        "description": "List of downloaded videos"
                    },
                    
                    
                },
                "description": "Video system",
                "type": "Video"
            }
}

Assistant:
The user needs to play a downloaded video, so I need to play the first video in the downloaded_videos list
```python
vw['video']['value']['current_video']['value'] = vw['video']['value']['downloaded_videos'][0]
```

### Example 2
User:
Turn up the volume a bit
Current system status:
vw = {
    "environment": {
                "value": {
                    "volume": {
                        "type": "int",
                        "value": 50,
                        "description": "Volume level (0-100)"
                    },
                    "sound_channel": {
                        "type": "str",
                        "value": "music",
                        "description": "
                                Current sound channel type, can be music, video, navigation, radio, conversation; the current environment has only one player, only one system can use the player at a time, so you need to modify the sound_channel field in the environment to use different systems.
                                - Please set to conversation when making/receiving calls or adjusting call volume
                                - Please set to music when playing music or adjusting music volume
                                - Please set to navigation when adjusting navigation volume, starting navigation, switching destinations, adding/removing waypoints, turning on/off announcements, or switching announcement modes
                                - Please set to radio when playing radio stations or adjusting radio volume
                                - Please set to video when adjusting video volume or playing videos
                                "
                    },
                    "unit_system": {
                        "type": "str",
                        "value": "mile",
                        "description": "Distance unit system, supports mile or kilometer"
                    }
                },
                "description": "World environment",
                "type": "type"
            }
}

Assistant:
The user needs to turn up the volume
```python
vw['environment']['value']['volume']['value'] = 60
```
Next, please generate state transition code to solve the user's request based on the current status of the system.
\end{lstlisting}

\begin{lstlisting}[
  style=mytext, 
  label={list:Plan},
  caption={prompt that guides agent to generate a plan first for reasoning experiment.},
  captionpos=t
]
You are an in-car AI assistant responsible for executing user requests. Based on the user's command, your first step is to generate a plan.

**Example Task Flow**
User:
Navigate to the nearest gas station and turn up the navigation volume a bit.

Assistant:
The user wants to start navigation to the nearest gas station and slightly increase the navigation volume. To fulfill this request, I need to identify and operate the modules responsible for navigation and volume control. I will also check which APIs are available in these modules to carry out the required actions.
\end{lstlisting}

\section{API Documentation}
\label{app:api_document}
\begin{longtable}{ll}
    \caption{API documentation for the Conversation module, detailing 15 phone and messaging functions with parameters, requirements, and value ranges for implementing voice communication capabilities.} \label{tab:conversation_api_summary} \\
    \toprule
    \multicolumn{2}{c}{\textbf{\texttt{conversation\_soundVolume\_increase}}} \\
    \midrule
    \textbf{Device} & conversation \\
    \textbf{Description} & Increase volume (0--100); use \texttt{value} or \texttt{degree} \\
    \textbf{Arguments} & \texttt{value} (\texttt{int}): numeric increase; exclusive with \texttt{degree} \\
                       & \texttt{degree} (\texttt{string}): \{"large", "little", "tiny"\} \\
    \textbf{Required} & None \\
    \midrule
    \multicolumn{2}{c}{\textbf{\texttt{conversation\_soundVolume\_decrease}}} \\
    \midrule
    \textbf{Device} & conversation \\
    \textbf{Description} & Decrease volume (0--100); use \texttt{value} or \texttt{degree} \\
    \textbf{Arguments} & \texttt{value} (\texttt{int}): numeric decrease; exclusive with \texttt{degree} \\
                       & \texttt{degree} (\texttt{string}): \{"large", "little", "tiny"\} \\
    \textbf{Required} & None \\
    \midrule
    \multicolumn{2}{c}{\textbf{\texttt{conversation\_soundVolume\_set}}} \\
    \midrule
    \textbf{Device} & conversation \\
    \textbf{Description} & Set volume (0--100); must provide either \texttt{value} or \texttt{degree} \\
    \textbf{Arguments} & \texttt{value} (\texttt{int}), \texttt{degree} (\texttt{string}): \{"max", "high", "medium", "low", "min"\} \\
    \textbf{Required} & One of: \texttt{value}, \texttt{degree} \\
    \midrule
    \multicolumn{2}{c}{\textbf{\texttt{conversation\_phone\_call}}} \\
    \midrule
    \textbf{Device} & conversation \\
    \textbf{Description} & Make a phone call \\
    \textbf{Arguments} & \texttt{contact} (\texttt{string}) \\
    \textbf{Required} & \{\texttt{contact}\} \\
    \midrule
    \multicolumn{2}{c}{\textbf{\texttt{conversation\_phone\_redial}}} \\
    \midrule
    \textbf{Device} & conversation \\
    \textbf{Description} & Redial phone \\
    \textbf{Arguments} & None \\
    \textbf{Required} & None \\
    \midrule
    \multicolumn{2}{c}{\textbf{\texttt{conversation\_phone\_answer}}} \\
    \midrule
    \textbf{Device} & conversation \\
    \textbf{Description} & Answer phone \\
    \textbf{Arguments} & None \\
    \textbf{Required} & None \\
    \midrule
    \multicolumn{2}{c}{\textbf{\texttt{conversation\_phone\_hangup}}} \\
    \midrule
    \textbf{Device} & conversation \\
    \textbf{Description} & Hang up phone \\
    \textbf{Arguments} & None \\
    \textbf{Required} & None \\
    \midrule
    \multicolumn{2}{c}{\textbf{\texttt{conversation\_message\_send}}} \\
    \midrule
    \textbf{Device} & conversation \\
    \textbf{Description} & Send SMS \\
    \textbf{Arguments} & \texttt{contact} (\texttt{string}), \texttt{content} (\texttt{string}) \\
    \textbf{Required} & \{\texttt{contact}\} \\
    \midrule
    \multicolumn{2}{c}{\textbf{\texttt{conversation\_message\_view}}} \\
    \midrule
    \textbf{Device} & conversation \\
    \textbf{Description} & View SMS \\
    \textbf{Arguments} & \texttt{contact} (\texttt{string}) \\
    \textbf{Required} & None \\
    \midrule
    \multicolumn{2}{c}{\textbf{\texttt{conversation\_contact\_view}}} \\
    \midrule
    \textbf{Device} & conversation \\
    \textbf{Description} & Find contact \\
    \textbf{Arguments} & \texttt{contact} (\texttt{string}) \\
    \textbf{Required} & \{\texttt{contact}\} \\
    \midrule
    \multicolumn{2}{c}{\textbf{\texttt{conversation\_call\_miss\_view}}} \\
    \midrule
    \textbf{Device} & conversation \\
    \textbf{Description} & View missed calls \\
    \textbf{Arguments} & None \\
    \textbf{Required} & None \\
    \midrule
    \multicolumn{2}{c}{\textbf{\texttt{conversation\_call\_record\_view}}} \\
    \midrule
    \textbf{Device} & conversation \\
    \textbf{Description} & View call history \\
    \textbf{Arguments} & None \\
    \textbf{Required} & None \\
    \midrule
    \multicolumn{2}{c}{\textbf{\texttt{conversation\_contact\_hag\_view}}} \\
    \midrule
    \textbf{Device} & conversation \\
    \textbf{Description} & Query user's contact list \\
    \textbf{Arguments} & None \\
    \textbf{Required} & None \\
    \midrule
    \multicolumn{2}{c}{\textbf{\texttt{conversation\_call\_handsFree\_switch}}} \\
    \midrule
    \textbf{Device} & conversation \\
    \textbf{Description} & Hands-free switch \\
    \textbf{Arguments} & \texttt{switch} (\texttt{boolean}) \\
    \textbf{Required} & \{\texttt{switch}\} \\
    \midrule
    \multicolumn{2}{c}{\textbf{\texttt{conversation\_contact\_delete}}} \\
    \midrule
    \textbf{Device} & conversation \\
    \textbf{Description} & Delete contact \\
    \textbf{Arguments} & \texttt{contact} (\texttt{string}) \\
    \textbf{Required} & \{\texttt{contact}\} \\
    \bottomrule
\end{longtable}

\section{Instruction to Participants}
\label{sec:instruction}
\label{app:expert-instruction}
These four tables provide comprehensive instructions for annotators. Table \ref{tab:api-design-instruction} guides API design assessment focusing on functionality and parameter structures. Table \ref{tab:test-set-instruction} directs test set construction with emphasis on authentic user interactions and scenario diversity. Table \ref{tab:function-call-evaluation} outlines criteria for evaluating function call outputs based on intent fulfillment and parameter accuracy. Table \ref{tab:annotation-purpose} contextualizes the annotation process, explaining how annotators' expert judgments establish benchmarks for comparing function call generation methodologies in automotive systems.
\begin{table}
\caption{Instruction for API design process.}
\label{tab:api-design-instruction}
\begin{tabular}{|p{\textwidth}|}
\hline
\textbf{Instruction:} \\
\hline
\textbf{1. Background:} We will provide you with API information. Each entry includes: \\
\begin{itemize}
\item The device using this API
\item Detailed description of the API functionality
\item Required parameter information
\end{itemize} \\
\hline
\textbf{2. Requirements Analysis Phase:} Carefully analyze the provided information, clarifying the following points: \\
\begin{itemize}
\item \textbf{Functional Boundaries:} Determine the core functionality and boundaries of the API, clarifying what tasks it needs to accomplish
\item \textbf{Use Cases:} Consider the scenarios and ways the API will be used in practical applications
\item \textbf{User Expectations:} Understand end-user expectations for this functionality and possible interaction patterns
\item \textbf{Device Constraints:} Analyze the characteristics and limitations of the target device, ensuring the API design matches the device capabilities
\end{itemize} \\
\hline
\textbf{3. API Design Principles:} Follow these general design principles: \\
\begin{itemize}
\item \textbf{Simplicity:} The API should be concise and clear, exposing only necessary functionality
\item \textbf{Consistency:} Maintain consistency in naming and structure for ease of understanding and use
\item \textbf{Intuitiveness:} Design parameters and return values to be intuitive, reducing the learning curve
\end{itemize} \\
\hline
\textbf{4. Structure Design:} Design the basic structure of the API, including: \\
\begin{itemize}
\item \textbf{Naming Conventions:} Design clear and intuitive API names based on functional characteristics
\item \textbf{Parameter Design:} Determine necessary input parameters and their data types
\item \textbf{Return Value Design:} Plan the response structure and content of the API
\end{itemize} \\
\hline
\textbf{5. Documentation:} Write clear and complete API documentation: \\
\begin{itemize}
\item \textbf{General Description:} Provide an overview and purpose of the API
\item \textbf{Parameter Details:} Explain in detail the purpose, type, format, and constraints of each parameter
\item \textbf{Call Examples:} Provide specific API call examples and response examples
\end{itemize} \\
\hline
\end{tabular}
\end{table}
\begin{table}
\caption{Instruction for Test Set Construction and Manual Screening.}
\label{tab:test-set-instruction}
\begin{tabular}{|p{\textwidth}|}
\hline
\textbf{Instruction:} \\
\hline
\textbf{1. Background:} We will provide you with user-intelligent cockpit interaction scenarios. Each scenario includes: 
\begin{itemize}
\item Environment initialization statement
\item User query statement
\item API call corresponding to the query statement
\end{itemize} \\
\hline
\textbf{2. Review and Screening Criteria:} Review each interaction scenario based on the following criteria: 
\begin{itemize}
\item \textbf{Authenticity:} Whether the scenario matches real users' in-vehicle interaction habits and expressions
\item \textbf{Clarity:} Whether the user query statement is clear and unambiguous
\item \textbf{Completeness:} Whether the scenario description contains sufficient contextual information to understand user intent
\item \textbf{Diversity:} Ensure the test set covers various interaction patterns and functional domains
\end{itemize} \\
\hline
\textbf{3. API Matching Assessment:} Evaluate the matching degree between query statements and API calls: 
\begin{itemize}
\item \textbf{Functional Match:} Whether the API call meets the functional requirements expressed in query
\item \textbf{Parameter Correctness:} Whether API parameters correctly reflect the specific requirements in query
\item \textbf{Contextual Relevance:} Whether the API call takes into account the contextual information of the scenario
\item \textbf{Edge Case Handling:} Whether the API response is reasonable for edge cases
\end{itemize} \\
\hline
\textbf{4. Scenario Annotation:} Add the following annotations to each scenario: 
\begin{itemize}
\item \textbf{Scenario Classification:} Label the functional category to which the scenario belongs (e.g., navigation, music control, environment control, etc.)
\item \textbf{Expression Type:} Label the type of user query expression (direct command, inquiry, ambiguous expression, etc.)
\item \textbf{Complexity Level:} Rate the complexity of the scenario (simple, medium, complex)
\end{itemize} \\
\hline
\textbf{Important Considerations:} \\
\begin{itemize}
\item Prioritize the authenticity and coverage of the test set, avoiding overly artificial expressions
\item Maintain diversity among different scenarios, avoiding excessive focus on specific functions
\item Pay special attention to edge cases and exception handling test scenarios
\item Ensure the test set includes user expressions of different complexities and language styles
\end{itemize} \\
\hline
\end{tabular}
\end{table}
\begin{table}
\caption{Human Expert Evaluation Guide for Function Call Outputs in Intelligent Cockpit Systems}
\label{tab:function-call-evaluation}
\begin{tabular}{|p{\textwidth}|}
\hline
\textbf{Instruction:} \\
\hline
\textbf{1. Background:} We will provide you with Function Call output samples from an intelligent cockpit system for evaluation. Each sample includes: \\
\begin{itemize}
\item The original user query expressing an intent or request (e.g., "adjust temperature to 22 degrees")
\item The Function Call (FC) output generated by the model, including API calls and parameters
\item Reference to available API specifications and documentation
\item The expected system state after function execution
\end{itemize} \\
\hline
\textbf{2. Review Function Call Output:} Carefully examine the model-generated function calls, paying close attention to: \\
\begin{itemize}
\item \textbf{Intent Fulfillment:} Does the function call sequence correctly address what the user requested?
\item \textbf{API Selection:} Are the appropriate APIs called for the requested task?
\item \textbf{Parameter Accuracy:} Are all parameter values correct and aligned with the user's request?
\item \textbf{Call Sequence:} Is the order of function calls logical and appropriate for achieving the task?
\end{itemize} \\
\hline
\textbf{3. Determine Correctness:} Based on your review, determine whether the Function Call output is correct or incorrect. \\
\begin{itemize}
\item A function call output is considered correct if it would result in the expected system state through valid API calls with appropriate parameters.
\end{itemize} \\
\hline
\textbf{4. Record Results:} For each Function Call output, record the following information: \\
\begin{itemize}
\item Your Assessment (Correct or Incorrect)
\item For incorrect assessments, note specific issues (wrong API, missing calls, incorrect parameters)
\end{itemize} \\
\hline
\textbf{Important Considerations:} \\
\begin{itemize}
\item Focus on practical outcomes and user intent fulfillment rather than superficial differences
\item Consider the context of an automotive environment when evaluating API appropriateness
\item If multiple valid approaches exist, consider the output correct if any valid approach is used
\item If you are unsure about correctness, consult with another expert evaluator
\end{itemize} \\
\hline
\end{tabular}
\end{table}
\begin{table}
\caption{Purpose and Application of Human Expert Annotations in Intelligent Cockpit Function Call Evaluation}
\label{tab:annotation-purpose}
\begin{tabular}{|p{\textwidth}|}
\hline
\textbf{Annotation Purpose and Research Methodology:} \\
\hline
\textbf{1. Research Objective:}
\begin{itemize}
\item The primary goal of this human expert annotation process is to establish a high-quality benchmark dataset for comparing different function call generation methodologies in intelligent cockpit systems.
\item These expert annotations serve as the gold standard against which both our proposed method and traditional approaches will be evaluated.
\item The central research question is: Which method produces function calls that more closely align with human expert judgments of correctness and appropriateness?
\end{itemize} \\
\hline
\textbf{2. Comparative Analysis Framework:} 
\begin{itemize}
\item \textbf{Baseline Comparison:} Human-annotated data will be used to assess how closely traditional function call methods match expert expectations versus our novel approach.
\item \textbf{Alignment Metrics:} We will quantify the degree of alignment between each automated method and human annotations using precision, recall, F1 scores, and custom alignment metrics.
\item \textbf{Error Pattern Analysis:} Discrepancies between both methods and human annotations will be categorized to identify systematic strengths and weaknesses of each approach.
\end{itemize} \\
\hline
\textbf{3. Expected Research Outcomes:} 
\begin{itemize}
\item \textbf{Method Validation:} Demonstrate whether our proposed method produces function calls that more accurately reflect human expert judgment compared to traditional approaches.
\item \textbf{Performance Gaps:} Identify specific scenarios or query types where the performance gap between methods is most significant.
\end{itemize} \\
\hline
\textbf{4. Practical Applications of Research Findings:} 
\begin{itemize}
\item \textbf{System Selection:} Determine which function call generation approach should be implemented in production intelligent cockpit systems.
\item \textbf{Hybrid Optimization:} Identify opportunities to combine strengths of both approaches based on comparison with human annotations.
\item \textbf{User Experience Enhancement:} Leverage insights from human annotations to improve the naturalness and reliability of in-vehicle voice command systems.
\end{itemize} \\
\hline
\textbf{5. Annotation Quality Control:} 
\begin{itemize}
\item Multiple expert annotators will evaluate each sample to ensure reliability and minimize individual bias.
\item Inter-annotator agreement metrics will be calculated to validate the consistency and quality of the human benchmark data.
\item The final gold standard will prioritize samples with high annotator consensus to ensure a reliable comparison baseline.
\end{itemize} \\
\hline
\end{tabular}
\end{table}

\end{document}